\pdfoutput=1
\documentclass[11pt]{article}

\usepackage[final]{acl}

\usepackage{times}
\usepackage{latexsym}

\usepackage[T1]{fontenc}
\usepackage[utf8]{inputenc}

\usepackage{microtype}

\usepackage{inconsolata}

\usepackage{graphicx}
\usepackage{booktabs}
\usepackage{multirow}
\usepackage{longtable}
\usepackage{pdflscape}
\usepackage{xcolor}
\usepackage{hyperref}
\hypersetup{colorlinks=true}
\usepackage{placeins}
\title{Representations of Fact, Fiction and Forecast in\\ Large Language Models: Epistemics and Attitudes}

\author{Meng Li, Michael Vrazitulis, David Schlangen\\
  University of Potsdam \\
  \texttt{\{meng.li, michael.vrazitulis,  david.schlangen\}@uni-potsdam.de}} 
\begin{document}
\maketitle
\begin{abstract}
Rational speakers are supposed to know what they know and what they do not know, and to generate expressions matching the strength of evidence. In contrast, it is still a challenge for current large language models to generate corresponding utterances based on the assessment of facts and confidence in an uncertain real-world environment. While it has recently become popular to estimate and calibrate confidence of LLMs with verbalized uncertainty, what is lacking is a careful examination of the linguistic knowledge of uncertainty encoded in the latent space of LLMs. In this paper, we draw on typological frameworks of epistemic expressions to evaluate LLMs' knowledge of epistemic modality, using controlled stories. Our experiments show that the performance of LLMs in generating epistemic expressions is limited and not robust, and hence the expressions of uncertainty generated by LLMs are not always reliable. To build uncertainty-aware LLMs, it is necessary to enrich the semantic knowledge of epistemic modality in LLMs.
\end{abstract}

% The lack of semantic knowledge of epistemic modality in LLMs may undermine the effort to build better alignment techniques. 
% The sense of reality in LLMs is greatly shaped by the sense of probability. In this research, we examine the problem of probability from semantic, epistemic and metaphysical perspectives, and show  the limitations of previous works on hullucination and uncertainty quantification. The linguistic encoding of fact and inference in LLMs is much less studied. Therefore, We develop a diagnostic dataset to test the linguistic knowledge of LLMs in using modal expressions. By evaluating the modal expressions of LLMs, we find interesting conclusions.

\section{Introduction}

\begin{figure}[ht!]
\centering
\includegraphics[width=0.45\textwidth]{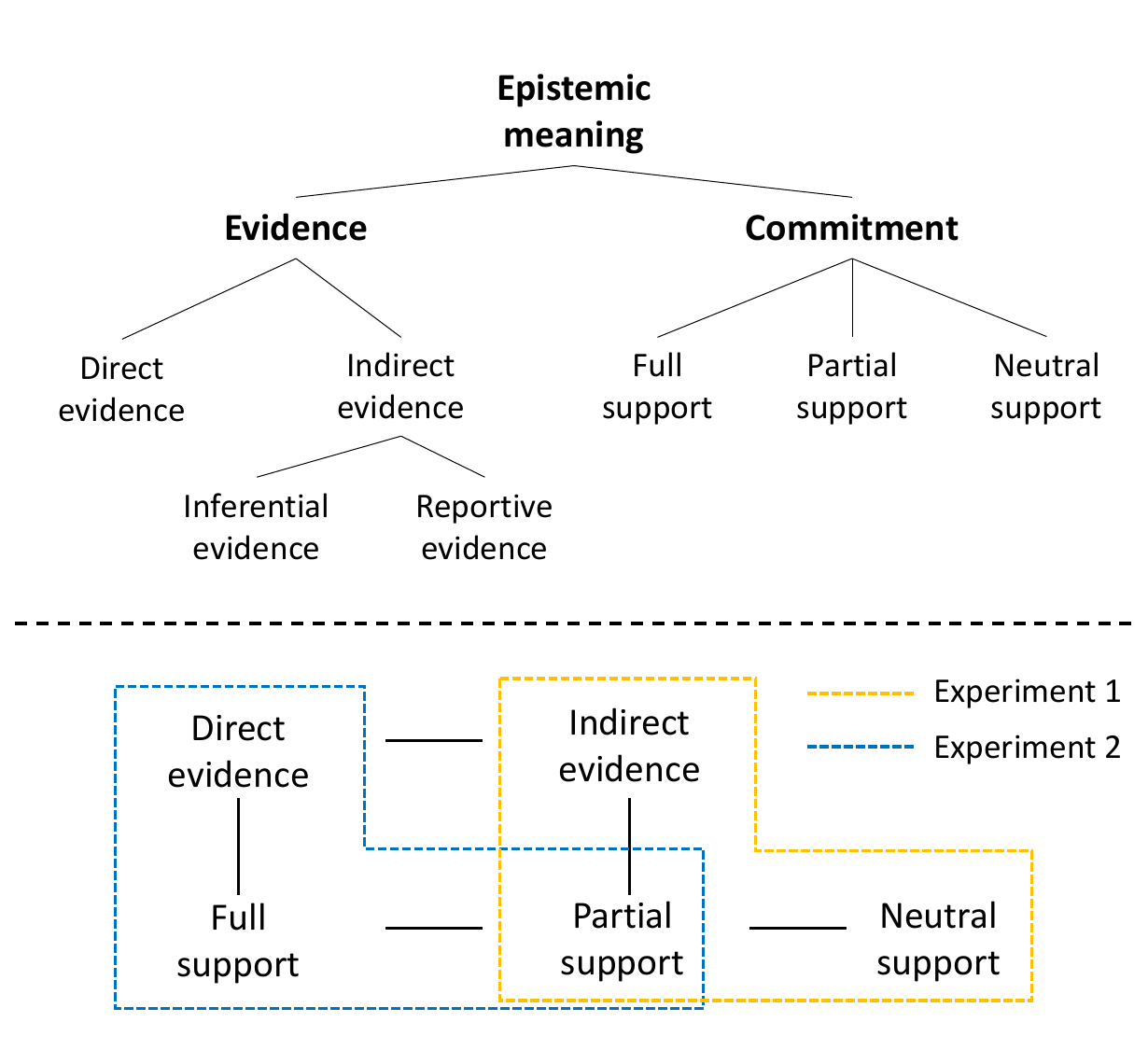}
\caption{Dimensions of epistemic meanings (above) and the semantic map of epistemic modality (below) \citep{boye2012epistemic}. The colored dashed lines indicate the mapping between concepts and our two experiments.}
\label{fig:sem-map}
\end{figure}

As LLMs are increasingly deployed in real-world situations, it is important for LLMs to behave in a rational way. They should be able to distinguish fact and imagination, and generate responses based on the credibility of the available evidence and the degree of their \textit{confidence}. For example, imagine that Tom and Alice are looking for a missing book. If Tom says, “The book \textit{may} be under the sofa”, he means that it is possible that the book is under the sofa according to the available evidence. Namely, the conclusion is compatible with existing evidence. If Tom says, “The book \textit{has to} be under the sofa”, he is indicating that he believes to have strong evidence for the conclusion that the book is underneath the sofa, and he is ruling out all other potential locations with Alice. Given the available evidence, these epistemic modal expressions encode knowledge of necessity and possibility. 

To build uncertainty-aware LLMs, many recent works focus on confidence estimation and calibration of LLMs. One common approach is about prompting language models to generate different levels of uncertainty in verbalized words or expressions \citep{mielke2022reducing, linteaching, xiongcan}. Given that current LLMs can generate very fluent responses, these papers assume, consciously or unconsciously, that these LLMs already master the linguistic knowledge of generating expressions of uncertainty, and the only problem left is alignment, namely how to elicit LLMs to generate these expressions at the right time. This assumption, however, was never carefully examined. Therefore, in this paper, we will investigate the form and meaning of epistemic expressions from typological perspectives, and evaluate the linguistic knowledge of epistemic modality present in an exemplary set of LLMs.

%“ ”

Let us first consider the meaning of epistemic expressions. The expressions of uncertainty in natural languages are semantically related to epistemic modality, because epistemic modality indicates the degree of certainty that a speaker has towards propositions \citep{portner2009modality, szarvas2012cross, giannakidou2021truth}. As shown in Figure~\ref{fig:sem-map}, epistemic meanings can be divided into \textit{evidence} and \textit{commitment} \citep{ boye2012epistemic}. Evidence is about the source of information or justification \citep{bybee1985morphology, palmer1986mood, aikhenvald2004evidentiality} and can be further divided into \textit{direct evidence} and \textit{indirect evidence}. Commitment is an epistemic modal meaning that expresses a speaker's degree of confidence or certainty \citep{Coates1983modal, bybee1994evolution, van1998modality}. It is conceived as a continuous, quantitative scale, which can be divided into three levels: \textit{full support}, \textit{partial support}, and \textit{neutral support} (see Appendix \ref{app:ling-background} for details). \footnote{To improve readability, we adapt terms in \citet{ boye2012epistemic} to more standard expressions in current practice across theories and frameworks. We use \textit{evidence} to replace \textit{epistemic justification} and use \textit{commitment} instead of \textit{epistemic support}.}

We now turn to the form of epistemic expressions. Epistemic meanings are expressed by various lexical, morphological, and syntactic structures across languages. In English, there are three major types of forms to express uncertainty: (1) modal adjectives and adverbs; (2) attitude verbs (or mental verbs); (3) modal auxiliaries and semi-auxiliaries. Modal adjectives and adverbs express probability through lexical meaning, such as \textit{certain}/\textit{certainly} and \textit{probable}/\textit{probably}. Attitude verbs, such as \textit{believe}, \textit{know} and \textit{doubt}, describe internal mental states towards propositions. Modal auxiliaries and semi-auxiliaries, like \textit{must}/\textit{may}/\textit{have to}, express modal meanings of necessity and possibility.

In previous work on confidence estimation and calibration, LLMs have been extensively evaluated on knowledge-intensive datasets such as TriviaQA \citep{joshi2017triviaqa}, StrategyQA \citep{geva2021did}, and GSM8K \citep{cobbe2021training}. However, these datasets present several limitations when it comes to evaluating LLMs' knowledge of epistemic modality. (1) Although these datasets provide the reference of factual correctness, the construct of knowledge-intensive tasks is to measure external world knowledge, not linguistic knowledge. They support the analysis of correspondence between the models' verbalized confidence and the likelihood that the answer is correct, but overlook the fact that human speakers also express their confidence in possible worlds, such as crime novels. (2) These datasets lack an agentic perspective, and they are coarse-grained. In other words, there are no controls on types of evidence and degrees of commitment. In addition, linguistic targets are also not structured systematically for comparison. (3) For large-scale real-world datasets, there is a risk of data contamination in the competitions for higher rankings in leaderboards. To address such concerns, we design controlled stories to test whether LLMs can predict the correct epistemic modal expressions (see Figure \ref{fig:exp1-design} and Figure \ref{fig:exp2-design}). By simplifying the complexity of reasoning and controlling different factors, we can see how prompt formats and modal semantics affect the generation of epistemic expressions. \footnote{Accessible data and code: \url{https://github.com/limengnlp/llm-fff}}

This paper offers the following key contributions: (1) As far as we know, this is the first paper to evaluate the linguistic knowledge of epistemic modality in LLMs, namely the underlying knowledge that allows LLMs to understand and generate epistemic expressions. We demonstrate the limited performance of LLMs in selecting appropriate epistemic expressions. (2) Through controlled experiments, we show how the number of parameters, prompt formats and modal semantics affect the accuracy of LLMs. For modal auxiliaries, LLMs have better performance for modal necessity than modal possibility. For attitude verbs, LLMs are better at reporting facts than beliefs under different degrees of epistemic certainty. (3) By relating our question to a typological framework which categorizes epistemic expressions, we disentangle \textit{linguistic} uncertainty from \textit{aleatoric} and \textit{epistemic} uncertainty \citep{kendall2017uncertainties}. We also offer insights on how to improve LLMs in the light of children's semantic development.

\section{Related Work}

\subsection{Teaching LLMs to Express Their Uncertainty Truthfully}

The problem of hallucinations and the need for alignment motivate confidence estimation and calibration in LLMs. There are different approaches for confidence estimation \citep{geng2024survey}: logit-based methods \citep{kuhnsemantic, huang2023look, vazhentsev2023efficient, duan2024shifting}, internal state-based methods \citep{ren2022out, kadavath2022language, burnsdiscovering, azaria2023internal, li2024inference}, verbalized methods \citep{mielke2022reducing, xiongcan}, consistency-based estimation \citep{manakul2023selfcheckgpt, lingenerating}, and surrogate methods \citep{shrivastava2023llamas}. Verbalized methods involve prompting language models to output different levels of uncertainty in words or numbers, and prove an effective approach to calibrating language models with verbalized metacognition. \citet{mielke2022reducing} train an external calibrator to guide language models to generate with appropriate levels of uncertainty. \citet{linteaching} fine-tune language models with a human annotated dataset to generate verbalized words and numbers at the same time. 

In addition, there is also research on the behavior of LLMs on the expressions of uncertainty. It was found empirically that LLMs are sensitive to the expressions of uncertainty injected in prompts, and these expressions can improve or impair their performance \citep{zhou2023navigating}. \citet{zhou-etal-2024-relying} report that LLMs are reluctant to express uncertainty when they generate wrong answers. \citet{yona-etal-2024-large} define a formal metric on faithful response uncertainty, and provide evidence that instruction-tuned LLMs perform poorly at conveying their intrinsic uncertainty. 

Of the several types of devices expressing uncertainty, adjectives and adverbs expressing probability received more attention, because it is easier to build the mapping between uncertainty expressions and explicit numerical responses from humans at the population level \citep{wallsten1986measuring, willems2019variability, fagenperception}. \citet{sileo-moens-2023-probing} show that neural language models struggle to understand words expressing probability, and fine-tuning with the supervision of human perception can lead to improvements. \citet{belem-etal-2024-perceptions} compare LLMs and humans in mapping uncertainty expressions to self-reported numerical probabilities.

\subsection{Theory of Mind and Language Models}
Theory of mind (ToM), the ability to infer other people's intents and beliefs, is assumed to play a key role in how children learn the meaning of words \citep{bloom2002children} and resolve reference ambiguities in conversations \citep{clark1981definite}. Thus, it is a crucial component of intelligent systems that interact with humans. Research shows that the acquisition of epistemic modality is related to the development of ToM abilities in children, because understanding the meaning of epistemic modality requires children's ability to handle representations of mental states \citep{gopnik1988children, papafragou2002modality}.

\citet{grant2017can} and \citet{nematzadeh2018evaluating} adapt ToM tests in developmental psychology to evaluate the ability of language models in question answering (QA) tasks. Existing ToM tests are usually generated with templated stories where there is information asymmetry in a limited set: ToM-bAbI \citep{nematzadeh2018evaluating},  ToMi \citep{le2019revisiting}, Hi-ToM \citep{wu2023hi}, OpenToM \citep{xu-etal-2024-opentom}. In recent years, there has been a surge of research on ToM behavior in LLMs \citep{sap2022neural, sclar2023minding, wilf2023think, ullman2023large, shapira2024clever, kosinski2024evaluating, strachan2024testing}. Our work leverages the templates of ToM stories and shifts the focus from inferring other agents' mental states to articulating one's own thinking truthfully. 

\subsection{Modal Semantics and LLMs} 
The semantics of conditionals are closely linked to modality. \citet{holliday2024conditional} test LLM reasoning in a set of inference patterns involving conditionals and epistemic modals, and identify the logical fallacies of LLMs. Our work aims to assess the linguistic knowledge of LLMs through their behavior in simple and controlled stories, without assuming a specific theory of logical or probabilistic reasoning. In other words, logical reasoning is not the focus here and its complexity is intentionally limited. 
% Grounding abstract words and probing their knowledge in LMs

\section{Method}

To understand what LLMs know, we assume that their knowledge cannot be directly measured, but can be inferred from observable behavior \citep{jiang2020can, hu2023prompting}.

\subsection {Grounding Epistemic Modality through Targeted Stories}  
% why use storyboards?
% paradigm in children language & field linguistics

%The central questions of an behavioral approach to evaluate the semantic knowledge of language models are as follows: (1) What observable behavior can be used as data? (2) How to collect or construct these data for a reliable analysis? (3) What analytical techniques are appropriate to link a dataset with a particular research question? We will answer the first and second questions in the context of modality. The third question will be answered in Sections \ref{sec: exp-1-analysis} and \ref{sec: exp-2-analysis}.

The knowledge of morphology and syntax encoded in language models can be observed directly from the generated text and their analysis is more explicit. However, meaning cannot be observed directly. To investigate the semantic knowledge of language models, it needs to be inferred from the behavior that such knowledge is assumed to underlie, especially responses to specific stimuli in experimentally controlled settings.

Compared with concrete words like visible objects and actions, the meaning of modal words is more abstract and highly context-dependent. To trigger participants or language models to generate modal expressions in a coherent and plausible way, we need contextualized stories. In the field of child language, the \textit{hidden object task} is designed to test children's understanding of epistemic modality \citep{hirst1982acquisition, noveck1996children, ozturk2015acquisition}. In \citet{ozturk2015acquisition}, children participants were presented with brief animated stories, where an animal would hide in a box. Given the information and prompt, participants were asked to reply with “Yes/No” to questions about the location of the animal. \textit{Prompt format}, \textit{modal semantics}, and \textit{types of stories} are controlled. In linguistic fieldwork, targeted storyboards also prove effective in studying modality \citep{burton2015targeted}. The key point is that the story is designed to include at least one targeted context, which can be utilized to test hypotheses about specific linguistic forms within that context. These practices provide transferrable research paradigms to evaluate the knowledge of epistemic modality in LLMs.
% Prompt format has two levels: statement prompt and question prompt. In the statement prompt, participants have to say “if they agree or not” after explanations from the experimenter. In the question prompt, participants just need to answer “Yes or No”. Modality is also splitted into two levels: possibility modal (\textit{may/can}; P-modal condition) and necessity modal (\textit{have to}; N-modal condition). 

There are different ways to generate stimuli for LLMs: (1) manually written datasets like GPQA \citep{rein2023gpqa}; (2) procedurally generated datasets such as BLiMP \citep{warstadt2020blimp}; (3) language model generated dataset like BigToM \citep{gandhi2024understanding}. In the present study, we generate stimuli by combining manual writing with template-based generation.

\begin{figure*}[ht!]
\centering
\includegraphics[width=1\textwidth]{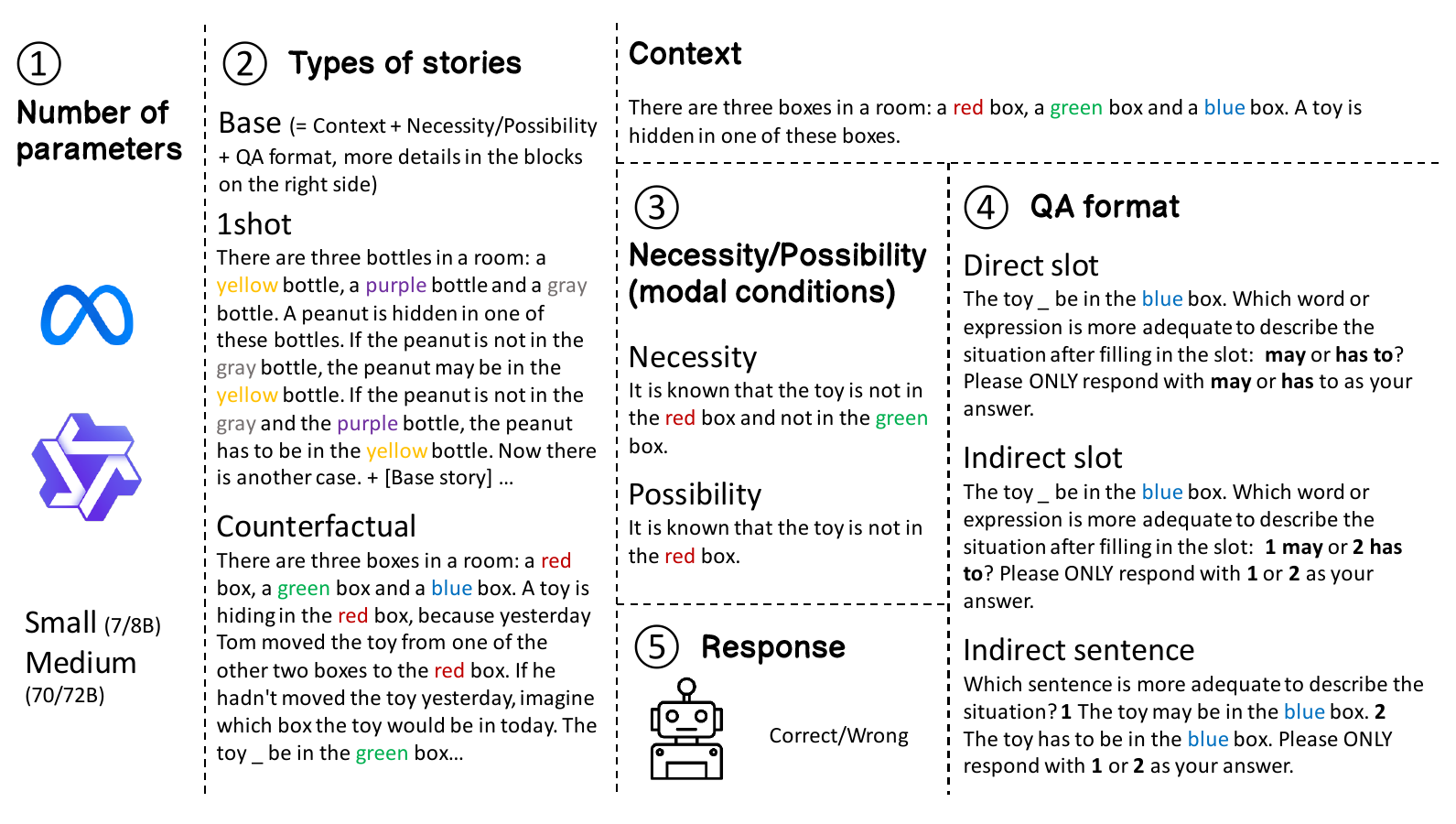}
\caption{Experimental design for assessing modal auxiliaries and semi-auxiliaries.}
\label{fig:exp1-design}
\end{figure*}

\subsection {Models} 
Considering the reproducibility of behavioral experiments, we evaluate eight open weight instruction-tuned models: \texttt{Llama3-8B/70B-Instruct}, \texttt{Llama3.1-8B/70B-Instruct} \citep{llama3modelcard}, \texttt{Qwen2-7B/72B-Instruct} \citep{qwen2}, \texttt{Qwen2.5-7B/72B-Instruct} \citep{qwen2.5}. They are accessed through the Huggingface Transformers library. We use greedy decoding for the result. These experiments were implemented on servers equipped with Nvidia A100 GPUs (80GB RAM).

\section{Experiment 1: Modal Auxiliaries}
% Purpose
% Participant
% Stimuli and Procedure

The first experiment systematically investigated LLMs' semantic knowledge of epistemic modal verbs, \textit{may}/\textit{might} vs \textit{must}/\textit{have to}, with a variety of modal scenarios. Eight instruction-tuned models were tested with simple stories that gave different cues about a set of elements and the selection of possibility/necessity. The number of parameters for these models falls into two categories: small (7-8B) and medium (70-72B). 

\subsection{Experimental Design}

There are 150 stories generated by five templates (see brief example in Figure \ref{fig:exp1-design} and more details in Appendix \ref{app:template}). In each story, there is a context to create a set of elements (objects, persons, places, etc.). In the condition with a necessity modal (\textit{N-modal condition}), there is clear and sufficient information to rule out all other candidates and identify the only one element left. In the condition with a possibility modal (\textit{P-modal condition}), the given information is not sufficient to make a precise statement, and there is uncertainty. For example, in a story generated with the \textit{hidden object} template, a toy is hidden in one of the three boxes: a red box, a green box, and a blue box. In the N-modal condition, the text provides enough information to rule out two boxes. If it is not in the red and green box, then it \textit{has to}/\textit{must} be in the blue box. In the P-modal condition, if the toy is not in the red box, it is still not certain that the toy is in the blue box or the green box. The toy \textit{may}/\textit{might} be in the blue box, which is more appropriate to describe this situation (see Figure~\ref{fig:exp1-design}).

We design three question and answer formats: direct slot, indirect slot, and indirect sentence, to ensure that responses to the modal statements reflected the semantic intuitions of LLMs. The direct slot format ask LLMs to select words that could be filled in a slot and to respond with these words directly. The indirect slot format asks LLMs to select words that could be filled in a slot, but to respond with the associated number or index. The indirect sentence format requires LLMs to select sentence-level statements and to respond with the associated number. The difference between slot and sentence formats is that LLMs need to identify the position of the slot, fill it with different words and compare. However, the sentence statements are much more natural to evaluate directly in the sentence format. The difference between direct and indirect formats is whether LLMs need metalinguistic indices (1/2) to refer epistemic modals when they respond. Theoretically, these indices increase the complexity of processing. If the knowledge of epistemic modal semantics in LLMs is robust, performance should be similar across different prompt formats. Otherwise, performance might diverge across the three different formats. 

% why different prompt format?
% the motivation of different types of stories (to be modified)

There are three different types of story: base, 1-shot and counterfactual stories. The base stories have been explained above. In-context learning (ICL) has proven to be an effective post-training technique to enable LLMs to solve new tasks with only a few demonstrations. One-shot stories introduce additional similar narratives to the base version, illustrating the selection of modal verbs in both N-modal and P-modal conditions. To prevent LLMs from simply copying the answer from the 1-shot example, we use lexical variations, such as differencing colors or person names. Modality concerns possible worlds and alternative ways that things could be. There are linguistic interactions between counterfactual conditions and epistemic modality. In English, counterfactual conditions are expressed through past tense and subjunctive mood. Therefore, we include counterfactual stories to create parallel possible scenarios by changing a verified true condition. These two variant types are designed to test whether the knowledge of epistemic modality in LLMs is sensitive to other information structures. Intuitively, for humans, the 1-shot stories should be easier than the base stories since there are additional supervised examples, while the counterfactual stories should be harder than the base stories, because they require additional counterfactual reasoning. Fifty stories were created for each type, resulting in a total of 150 stories. %And these stories were organized in a ﬁxed order.

We measure the performance with \textit{accuracy} and \textit{paired accuracy}. For paired accuracy, we count the answer as correct if the questions in a pair of N-modal and P-modal conditions are both answered correctly.

\subsection{Statistical Analysis and Results}
\label{sec: exp-1-analysis}

\begin{figure}[ht!]
\centering
\includegraphics[width=0.48\textwidth]{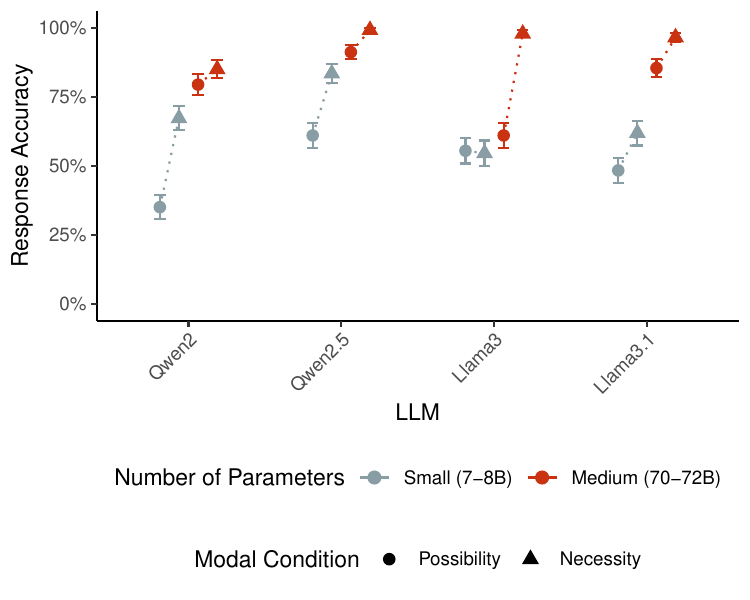}
\caption{Response accuracy in Experiment~1 by LLM, number of parameters, and modal condition. Error bars represent 95\% confidence intervals.}
\label{fig:exp1-acc-dotplot-np}
\end{figure}

\begin{table}[h]
\small
\centering
\begin{tabular}{lrr}
\toprule
\textbf{LLM} & \textbf{Acc} & \textbf{Paired Acc} \\
\midrule
Qwen2-7B        & 51.2 & 2.4 \\
Qwen2.5-7B      & 72.3 & 45.1 \\
Llama3-8B       & 55.1 & 11.8 \\
Llama3.1-8B    & 55.2 & 11.8 \\
\midrule
Qwen2-72B      & 82.3 & 64.7  \\
Qwen2.5-72B     & 95.3 & 90.7  \\
Llama3-70B      & 79.6 & 59.1  \\
Llama3.1-70B    & 91.1 & 82.2  \\
\bottomrule
\end{tabular}
\caption{Mean accuracy and mean paired accuracy in percent, for different LLMs in Experiment~1.} 
\label{table: exp1-acc}
\end{table}

Mean accuracies and paired accuracies for all assessed LLMs are reported in Table~\ref{table: exp1-acc}. Judging from these descriptive statistics alone, it seems that LLMs with a medium number of parameters (70-72B), with accuracies ranging from 79.6\% to 95.3\%, clearly outperform their counterparts with a smaller number of parameters (7-8B), whose accuracies just range between 55.1\% and 72.3\%. 

Using logistic regression, we assess how number of parameters, story type, modal condition, and QA format possibly modulate response accuracy. To do so, we fit a separate logistic regression model to each of the four classes of LLMs (Qwen2, Qwen2.5, Llama3, Llama3.1). Details on how these statistical analyses were performed are reported in Appendix~\ref{app:logistic:exp1}.

Across all assessed LLMs, the number of parameters has a significant impact on response accuracy (Qwen2:\ \emph{b}\,=\,1.63, \emph{SE}\,=\,0.12, \emph{p}\,<\,.001; Qwen2.5:\ \emph{b}\,=\,3.15, \emph{SE}\,=\,0.36, \emph{p}\,<\,.001; Llama3:\ \emph{b}\,=\,2.17, \emph{SE}\,=\,0.20, \emph{p}\,<\,.001; Llama3.1:\ \emph{b}\,=\,2.96, \emph{SE}\,=\,0.22, \emph{p}\,<\,.001), indicating that a large number of parameters (70--72B) leads to increased accuracy. A~consistent effect of modal condition is also found across LLMs (Qwen2:\ \emph{b}\,=\,0.87, \emph{SE}\,=\,0.12, \emph{p}\,<\,.001; Qwen2.5:\ \emph{b}\,=\,1.99, \emph{SE}\,=\,0.32, \emph{p}\,<\,.001; Llama3:\ \emph{b}\,=\,1.79, \emph{SE}\,=\,0.19, \emph{p}\,<\,.001; Llama3.1:\ \emph{b}\,=\,1.14, \emph{SE}\,=\,0.17, \emph{p}\,<\,.001). The positive sign of the effect indexes that accuracy was higher for Necessity trials than it was for Possibility trials. The interaction between number of parameters and modal condition is also significant across LLMs, yet the sign and magnitude of the interaction differs considerably between LLMs (Qwen2:\ \emph{b}\,=\,$-$0.93, \emph{SE}\,=\,0.23, \emph{p}\,<\,.001; Qwen2.5:\ \emph{b}\,=\,1.54, \emph{SE}\,=\,0.63, \emph{p}\,=\,.015; Llama3:\ \emph{b}\,=\,3.66, \emph{SE}\,=\,0.38, \emph{p}\,<\,.001; Llama3.1:\ \emph{b}\,=\,1.16, \emph{SE}\,=\,0.33, \emph{p}\,<\,.001). Figure~\ref{fig:exp1-acc-dotplot-np} illustrates the effects of number of parameters and modal condition.
% \hyperref[fig:exp1-acc-dotplot-np]{Figure~\ref{fig:exp2-design}}.

\begin{figure*}[ht!]
\centering
\includegraphics[width=1\textwidth]{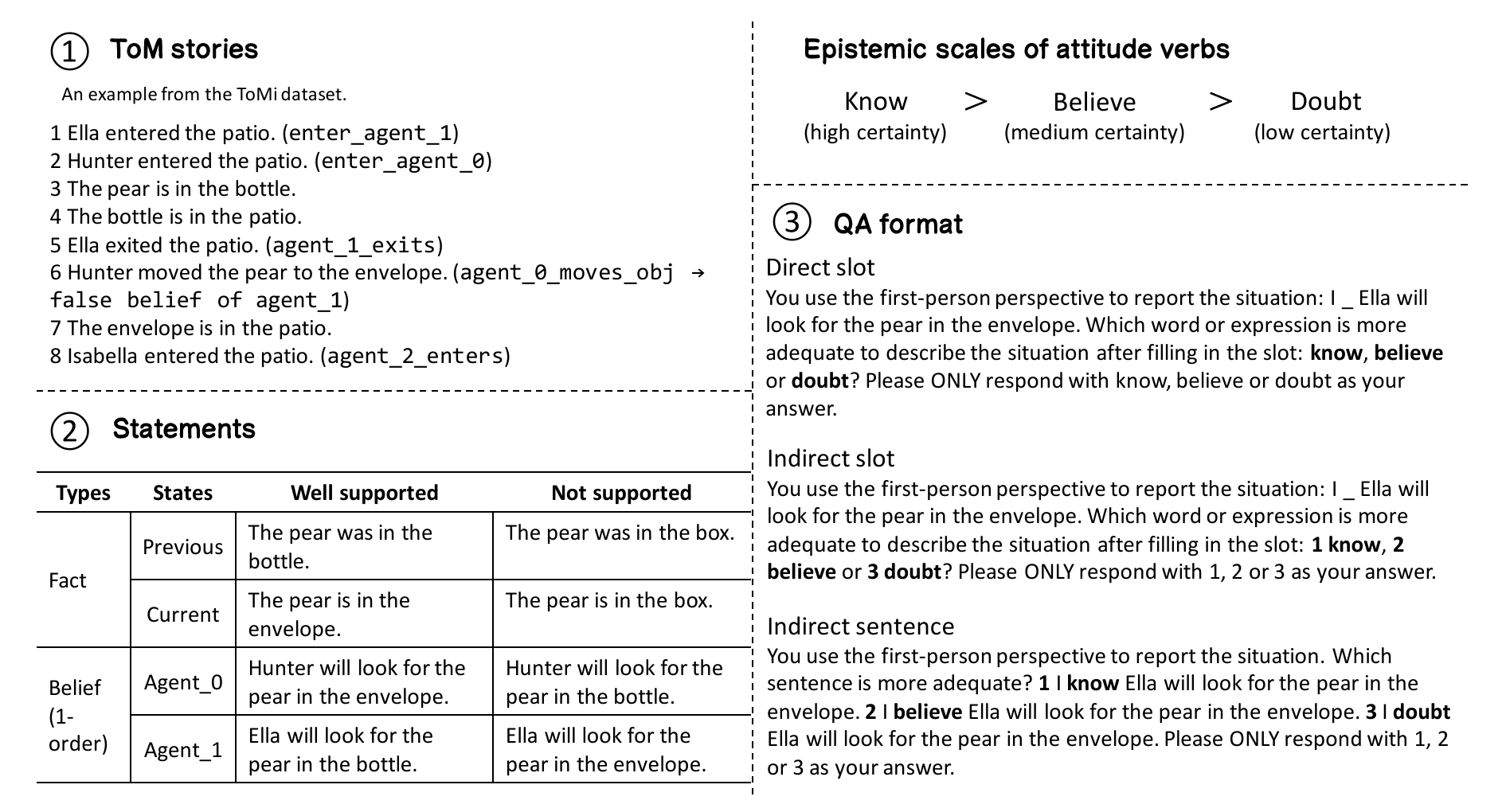}
\caption{Stimuli in experiment 2 to assess attitude verbs.}
\label{fig:exp2-design}
\end{figure*}

The factors story type and QA format do also modulate response accuracy to some extent. However, the magnitudes of the effects of story type and QA format (as well as their interactions with number of parameters) are rather small, and none of their effect signatures are consistent across the assessed LLMs.

%For example, the comparison between the base story type and the counterfactual story type yields either no effect, an effect favoring the base story type, or an effect favoring the counterfactual story type, depending on the LLM (Qwen2:\ \emph{b}\,=\,$-$0.50, \emph{SE}\,=\,0.16, \emph{z}\,=\,$-$3.15, \emph{p}\,=\,.002; Qwen2.5:\ \emph{b}\,=\,0.81, \emph{SE}\,=\,0.39, \emph{z}\,=\,2.09, \emph{p}\,=\,.036; Llama3:\ \emph{b}\,=\,$-$0.01, \emph{SE}\,=\,0.16, \emph{z}\,=\,$-$0.04, \emph{p}\,=\,.971; Llama3.1:\ \emph{b}~=~$-$0.52, \emph{SE}~=~0.19, \emph{z}~=~$-$2.67, \emph{p}\,=\,.007). For full reference of all corresponding regression estimates, see Tables \ref{table: exp1-Qwen2}, \ref{table: exp1-Qwen2.5}, \ref{table: exp1-Llama3}, \ref{table: exp1-Llama3.1} in Appendix~\ref{app:statistical:exp1} and Figures \ref{fig:exp1-acc-dotplot-exp}, \ref{fig:exp1-acc-dotplot-qa} in Appendix~\ref{app‌:figures:exp1}.

In sum, there is a salient improvement in accuracy when the expected correct response expresses a necessity, rather than a possibility. Further, as expected, performance is increased substantially when LLMs have a medium (70--72B) as opposed to just a small (7--8B) number of parameters. The higher accuracy for necessity modals suggests that models handle contexts with a unique, unambiguous conclusion more reliably than those requiring reasoning under uncertainty. For instance, when two out of three locations are ruled out, models correctly infer ``The toy must be in the blue box.'' In contrast, when multiple outcomes remain plausible, models often fail to select ``may'', indicating a weaker grasp of possibility in uncertain contexts.

\section{Experiment 2: Attitude Verbs}
% Purpose
% Participant
% Stimuli and Procedure
The second experiment focuses on the semantic knowledge of attitude verbs in LLMs:  \textit{know}, \textit{believe}, and \textit{doubt}. Given Theory-of-Mind (ToM) stories, LLMs were required to select different attitude verbs to report facts or beliefs with different degrees of certainty.

\subsection{Experimental Design}

% ToM design; meta-cognitive underpinning;
Self-concept (awareness of one's own mental states and functions) and a theory of mind (awareness of others' mental states and functions) are components of metacognition. The metacognitive system involves a second-order form of \textit{knowing} and can be viewed as an interface between the mind and reality \citep{demetriou2010development}. While ToM datasets are typically used to evaluate LLMs' ability to reason about others' beliefs in simple narratives, they can also be repurposed to assess whether LLMs can truthfully describe their own reasoning, given a shift in the focus of the task.

In a typical ToM test, the test subject (a child or a language model) observes a sequence of actions of two agents: \texttt{agent 0} moves an object into a container, and \texttt{agent 1} moves the object to another container when \texttt{agent 0} is not aware of this. The test subject is then asked questions about the actual state of the world and the agents' beliefs. In previous ToM datasets, the task is designed as a QA task, and questions were designed to ask for the location of the object in different settings. Here we change the questions and test the knowledge of LLMs in selecting attitude verbs (\textit{know}, \textit{believe}, and \textit{doubt}) to report facts or beliefs in different statements. To answer the question correctly, the model needs to choose the right verb based on the type of statement (fact/belief) and the strength of evidence. \footnote{Negation will change the values of an epistemic scale \citep{Horn1989-HORANH} and introduce more complex inference. For example, I am \textit{certain that not P} vs. I am \textit{not certain that P}. To keep the setting simple and focus on attitude verbs, we exclude negation here.}

Thirty stories were selected randomly from the ToMi dataset \citep{le2019revisiting, sclar2023minding}. For each story, we constructed eight statements: two statements on previous facts; two statements on current facts, two statements on 1-order beliefs of \texttt{agent 0}, and two statements on 1-order beliefs of \texttt{agent 1}. All paired statements in these four subtypes are contrastive in low/not low certainty (see examples in Figure \ref{fig:exp2-design}). In addition, there are also three question and answer formats: direct slot, indirect slot, and indirect sentence, which is similar to Experiment~1.

We measure the performance with \textit{accuracy}, \textit{paired accuracy}, and \textit{joint accuracy}. For paired accuracy, we count the answer as correct if questions in a pair of low and not low certainty conditions are both answered correctly, since these prompts are constructed in a controlled way. For joint accuracy, we count the answer as correct if questions about eight statements of the same ToM story are all answered correctly.

\subsection{Statistical Analysis and Results}
\label{sec: exp-2-analysis}

\begin{figure}[ht!]
\centering
\includegraphics[width=0.48\textwidth]{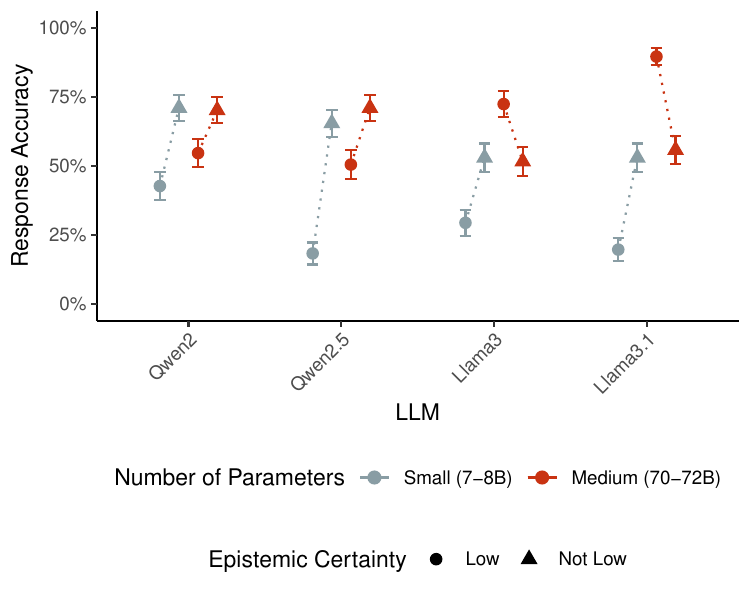}
\caption{Response accuracy in Experiment 2 by LLM, number of parameters, and modal condition. Error bars represent 95\% confidence intervals.}
\label{fig:exp2-acc-dotplot-certainty}
\end{figure}

% add joint accuracy

\begin{table}[h]
\small
\centering
\begin{tabular}{lrrr}
\toprule
\textbf{LLM} & \textbf{Acc} & \textbf{Paired Acc} & \textbf{Joint Acc}\\
\midrule
Qwen2-7B        & 56.9 & 38.6 & 0\\
Qwen2.5-7B      & 41.9 & 16.1 & 0\\
Llama3-8B       & 41.3 & 28.9 & 0\\
Llama3.1-8B    & 36.4 & 12.2 & 0\\
\midrule
Qwen2-72B      & 62.5 & 42.2 & 4.4\\
Qwen2.5-72B     & 60.8 & 46.4 & 5.6 \\
Llama3-70B      & 62.1 & 40.3 & 0\\
Llama3.1-70B    & 72.8 & 54.2 & 0\\
\bottomrule
\end{tabular}
\caption{Mean accuracy, paired accuracy, and joint accuracy in percent, for different LLMs in Experiment~2.}
\label{table: exp2-acc}
\end{table}

Table~\ref{table: exp2-acc} reports mean accuracies and paired accuracies for the assessed LLMs in Experiment~2. Again, LLMs with a medium number of parameters (70--72B) perform noticeably better than their counterparts with a small number of parameters (7--8B), by about 20 percentage points in accuracy (60.8--72.8\% vs.\ 36.4--56.9\%).

We use logistic regression to statistically model the effects of different factors on response accuracy. Specifically, we are interested in whether an LLM's number of parameters, the epistemic certainty of the attitude verb (low for \emph{doubt}, not low for \emph{believe} and \emph{know}), the type of statement (belief- or fact-based), and finally the QA format (direct/indirect slot, indirect sentence) each modulate the response accuracies on the ToM task in some meaningful way. Appendix~\ref{app:logistic:exp2} provides more details on how these statistical analyses were performed.

For Qwen2, the number of parameters does not show an effect on response accuracy in the ToM task (\emph{b}\,=\,0.08, \emph{SE}\,=\,0.15, \emph{p}\,=\,.604). Yet, for each of the remaining three LLM classes, an increase in the number of parameters (from small to medium) does in fact lead to a significant increase in response accuracy (Qwen2.5:\ \emph{b}\,=\,1.65, \emph{SE}\,=\,0.18, \emph{p}\,<\,.001; Llama3:\ \emph{b}\,=\,1.88, \emph{SE}\,=\,0.19, \emph{p}\,<\,.001; Llama3.1:\ \emph{b}\,=\,2.70, \emph{SE}\,=\,0.19, \emph{p}\,<\,.001).

Figure~\ref{fig:exp2-acc-dotplot-certainty} shows that, for all LLMs with a small number of parameters, target verbs with a relatively high epistemic certainty (\emph{believe} or \emph{know}) are associated with higher response accuracies, compared to a target verb with low epistemic certainty (\emph{doubt}). This trend also holds for the Qwen2/Qwen2.5 models with a medium number of parameters, but is interestingly reversed for the Llama3/Llama3.1 models with a medium number of parameters. These effect patterns are also reflected in the regression estimates for the main effect of epistemic certainty (Qwen2:\ \emph{b}\,=\,1.64, \emph{SE}\,=\,0.16, \emph{p}\,<\,.001; Qwen2.5:\ \emph{b}\,=\,3.08, \emph{SE}\,=\,0.22, \emph{p}\,<\,.001; Llama3:\ \emph{b}\,=\,1.36, \emph{SE}\,=\,0.32, \emph{p}\,<\,.001; Llama3.1: \emph{b}\,=\,$-$0.55, \emph{SE}\,=\,0.17, \emph{p}\,=\,.001) and the interaction between number of parameters and epistemic certainty (Qwen2:\ \emph{b}\,=\,$-$1.61, \emph{SE}\,=\,0.33, \emph{p}\,<\,.001; Qwen2.5:\ \emph{b}\,=\,$-$2.99, \emph{SE}\,=\,0.44, \emph{p}\,<\,.001; Llama3:\ \emph{b}\,=\,$-$5.30, \emph{SE}\,=\,0.64, \emph{p}\,<\,.001; Llama3.1: \emph{b}\,=\,$-$4.78, \emph{SE}\,=\,0.34, \emph{p}\,<\,.001).

The main effect of statement type (belief- vs.\ fact-based) is large and highly significant across LLM classes (Qwen2:\ \emph{b}\,=\,3.02, \emph{SE}\,=\,0.17, \emph{p}\,<\,.001; Qwen2.5:\ \emph{b}\,=\,3.79, \emph{SE}\,=\,0.22, \emph{p}\,<\,.001; Llama3:\ \emph{b}\,=\,4.79, \emph{SE}\,=\,0.34, \emph{p}\,<\,.001; Llama3.1: \emph{b}\,=\,2.74, \emph{SE}\,=\,0.19, \emph{p}\,<\,.001), indicating substantially higher response accuracies for fact-based statements. Figure~\ref{fig:exp2-acc-dotplot-statement} in Appendix~\ref{app‌:figures:exp2} confirms this visually.

Other included predictors fail to show consistent and salient effects across the assessed LLMs.

To summarize, the logistic regression analyses allow for three key observations about the assessed LLMs' behavior on the ToM task: (1) LLMs with a medium (rather than small) number of parameters tend to respond more accurately. (2) Relatively high epistemic certainty (\emph{believe} or \emph{know}) leads to higher accuracy, but note that the Llama3/Llama3.1 models with a medium number of parameters display the opposite effect. (3) LLMs systematically show higher accuracies on fact-based statements than on belief-based statements.

\section{Discussion}

\subsection{Comparing the Behavior of Epistemic Modality in Human and Machines}

\paragraph{Epistemic modal verbs} \citet{ozturk2015acquisition} tested children with the \textit{hidden object task} and evaluated their knowledge of modal expressions with \textit{may} and \textit{have to}. It is reported that children between the ages of 4 and 5 years have a basic understanding of epistemic semantic modals, but their knowledge of epistemic modals is sensitive to the syntactic--semantic context (statement vs.\ question prompts). Specifically, children were better at evaluating statements than at answering questions. Similarly, the performance of LLMs is also affected by prompt formats (see Figure \ref{fig:exp1-acc-dotplot-qa}). As a control group, adults achieve 97-100\% accuracy in different stories. By contrast, there is still a certain gap between the performance of LLMs and adults.

\paragraph{Attitude verbs} Different from words referring to concrete objects or visible actions, attitude verbs describe internal states of mind and leave few cues in the physical world. These characteristics pose challenges for native language learners. It is not until well into preschool that children begin to show adult-like performance on tasks involving attitude verbs \citep{hacquard2022acquisition}. There are different hypotheses and theories about the learning of attitude verbs \citep{landau1985language, gleitman1990structural, diessel2001acquisition, montgomery2002mental, papafragou2007we, israel2008mental, becker2013harder, hacquard2019children}. Such accounts, in turn, can also raise important questions for scientists in different research communities: 
\begin{itemize}
    \item For researchers of machine learning and computational linguistics, who are more interested in how LLMs learn and use attitude or mental verbs, the potential questions include: (1) Can a language model learn the meaning of attitude verbs without consciousness? (2) Are existing LLM training pipelines, such as self-supervised pretraining + supervised fine-tuning (SFT) + reinforcement learning from human feedback (RLHF), sufficient to learn the meaning of attitude verbs? (3) How do LLMs benefit from theories of children's semantic development (e.g., learning from interactions in pragmatically informative environments)?
    \item For cognitive scientists who specialize in developmental psycholinguistics and language acquisition, LLM's training mechanism may also provide a computational modeling perspective to validate or challenge existing learning theories of attitude verbs. For example, LLM pretraining employs self-supervised learning and does not explicitly use a priori syntactic categories. Does this imply that the cues of statistical distribution contribute to learning and that a priori syntactic categories are not necessary?
\end{itemize}
% summarize and compare the acquisition of modal auxiliaries
% bootstrapping theory of acquisition of attitude verbs for children

\subsection{Future Work}

% other linguistic devices in under-resourced languages
\paragraph{Forms of Modality in Low-Resource Languages} There are different means to express modal semantics in non-English languages: modal affixes, modal case, etc.\ \citep{de2006typological}. These morphological variations in low-resource languages impose challenges for building truthful multilingual LLMs, and remain to be tested in the future.

% connections with evidentials
% explanation of the framework| (1) direct:text -> multimodal; (2) indirect:conflicting evidences;
\paragraph{Enriching Benchmarks with Multimodal Evidence and Complex Reasoning} (1) We used text-based stories in both experiments, but future work could test epistemic reasoning in multimodal environments, especially as embodied intelligence becomes increasingly prominent. (2) In order to avoid interference from unnecessary world knowledge, we controlled the complexity of reasoning. However, LLMs still need to improve how they handle conflicting evidence \citep{kazemi2023boardgameqa, wan-etal-2024-evidence}.

\section{Conclusion}
In this paper, we evaluate the semantic knowledge of epistemic modality in open-weights LLMs through controlled stories, and show their limited performance in generating appropriate epistemic expressions. This implies that responses containing epistemic uncertainty from LLMs may be unreliable. Insufficient semantic knowledge of epistemic modality is a potential reason why LLMs are not good at truthfully expressing uncertainty. Therefore, to build rational LLMs, we should not only improve the methods of uncertainty estimation and calibration, but also enrich the semantic representation of epistemic modality.

\section*{Acknowledgments}
We are grateful to Jiangtian Li and the anonymous reviewers for helpful feedback.

\section*{Limitations}

We follow the behavioral approach to evaluate the semantic knowledge of epistemic modality in LLMs, and there are several limitations in this work. (1) Children's acquisition profiles and data from adult groups in existing literature on semantic development \citep{noveck1996children, ozturk2015acquisition, hacquard2022acquisition} provide indirect evidence that adults can achieve high performance on the tasks in this paper. However, we did not test human participants directly. (2) We did not leverage logit probabilities to design new metrics of intrinsic uncertainty, nor did we build mapping between model logits and self-reported human responses. (3) Our work focuses on the English language, and we did not study whether subword tokenization in LLMs can handle the morphological encoding of modality in low-resource languages.

% Points to mention:
% didn't test humans
% experiments didn't include complex stories
% didn't check logit information
% non-English language
% etc

%\bibliography{anthology, custom}
\bibliography{custom}

\appendix
%\onecolumn
%\twocolumn
%\clearpage
%\newpage
\section{More Background on Epistemic Meaning}
\label{app:ling-background}
Evidence is also often referred to by the terms \textit{evidentiality} \citep{aikhenvald2004evidentiality} or \textit{evidentials} \citep{bybee1994evolution}. Direct evidence covers “firsthand/visual/auditory/participatory evidence”. Within indirect evidence, \textit{reportive} and \textit{inferential evidence} can be distinguished from one another: reportive evidence is about “second-hand/hearsay/quotative evidence”; inferential evidence covers “inferential/assumptive evidence”. 

Full support has the strongest strength on the epistemic modal scale, and is equivalent to “knowledge/certainty (that not)/ 
epistemic impossibility”. Neutral support covers “epistemic possibility/(complete) uncertainty”. Partial support lies between full support and neutral support, and covers “probability/(relative weak) uncertainty/(un)likely/epistemic necessity”. 

A \textit{semantic map} is a visual representation of cross-linguistic patterns or regularities in semantic structures, which maps how various languages categorize meaning in a specific domain. The categories of evidence and commitment constitutes a continuous region in the semantic map of epistemic expressions (see Figure \ref{fig:sem-map}). We use nodes in the map to show the relation of two experiments in this paper.

%\clearpage 
%\twocolumn
\section{Contrast Coding for Experiments}
\label{app:contrast}

\subsection{Experiment 1}
\label{app:contrast:exp1}

\begin{table}[h]
\centering
\small
\begin{tabular}{lrr}
\toprule
\multicolumn{2}{l}{\textbf{Number of Parameters}} & Medium \\
\midrule
Small & & $-$0.5\\
Medium & & 0.5 \\
\midrule
\textbf{Modal Condition} & & Necessity \\
\midrule
Possibility & & $-$0.5\\
Necessity & & 0.5 \\
\midrule
\textbf{Story Type} & 1-Shot & Counterfactual \\
\midrule
Base & $-$0.5 & $-$0.5 \\
1-Shot & 0.5 & 0 \\
Counterfactual & 0 & 0.5 \\
\midrule
\textbf{QA Format} & Ind.~Slot & Ind.~Sent \\
\midrule
Direct Slot & $-$0.5 & $-$0.5 \\
Indirect Slot & 0.5 & 0 \\
Indirect Sentence & 0 & 0.5 \\
\bottomrule
\end{tabular}
\caption{Contrast coding for statistical analyses of Experiment~1. The factors are number of parameters, story type, modal condition, and QA format.} 
\label{table: exp1-cod}
\end{table}
%The factors are number of parameters, story type, modal condition, and QA format.

\subsection{Experiment 2}
\label{app‌:contrast:exp2}

\begin{table}[ht!]
\centering
\small
\begin{tabular}{lrrr}
\toprule
\multicolumn{2}{l}{\textbf{Number of Parameters}} & & Medium \\
\midrule
Small & & & $-$0.5\\
Medium & & & 0.5 \\
\midrule
\textbf{Epistemic Certainty} & & & Not Low \\
\midrule
Low & & & $-$0.5\\
Not Low & & & 0.5 \\
\midrule
\textbf{Statement Type} & Fact & Agent 1 & Current \\
\midrule
Belief (Agent 0) & $-$0.5 & $-$0.5 & 0 \\
Belief (Agent 1) & $-$0.5 & 0.5 & 0 \\
Fact (Previous) & 0.5 & 0 & $-$0.5 \\
Fact (Current) & 0.5 & 0 & 0.5 \\
\midrule
\textbf{QA Format} & & Ind.~Slot & Ind.~Sent \\
\midrule
Direct Slot & & $-$0.5 & $-$0.5 \\
Indirect Slot & & 0.5 & 0 \\
Indirect Sentence &  & 0 & 0.5 \\
\bottomrule
\end{tabular}
\caption{Contrast coding for statistical analyses of Experiment 2.} 
\label{table: exp2-cod}
\end{table}
%The factors are number of parameters, story type, modal condition, and QA format.

\section{Additional Information on Logistic Regression Analyses}
\label{app:logistic}

\subsection{Experiment 1}
\label{app:logistic:exp1}

We consider all main effects of the factors number of parameters, story type, modal condition, and QA format, in addition to each two-way interaction between number of parameters and any of the remaining factors. We focus specifically on the interactions between number of parameters and any other factor because it is plausible to assume that at scale (i.e., with more parameters) LLMs display clear qualitative differences in their response patterns. This, in turn, may modulate the effect patterns associated with any of the remaining factors.

In order to probe the relevance of potential random effects, we applied forward model selection based on the Akaike Information Criterion (AIC; \citealp{Akaike1973}), as implemented in the \emph{buildmer} package (\citealp{Voeten2024}) for the R programming language (\citealp{R2024}). We assessed random intercepts and random slopes for number of parameters, varying by LLM, by template, and by item nested within template. However, the model selection indicated that none of the assessed random effect terms substantially improved goodness of model fit as measured by the AIC. That is why eventually we fitted simple logistic regression models, without any random effects, whose results we report below.

In some cases, AIC-based model selection led to dropping certain fixed-effect interaction terms, as their inclusion did not improve the overall goodness of fit enough to offset the AIC penalty for additional parameters.

\begin{figure}[ht!]
\centering
\includegraphics[width=0.9\textwidth]{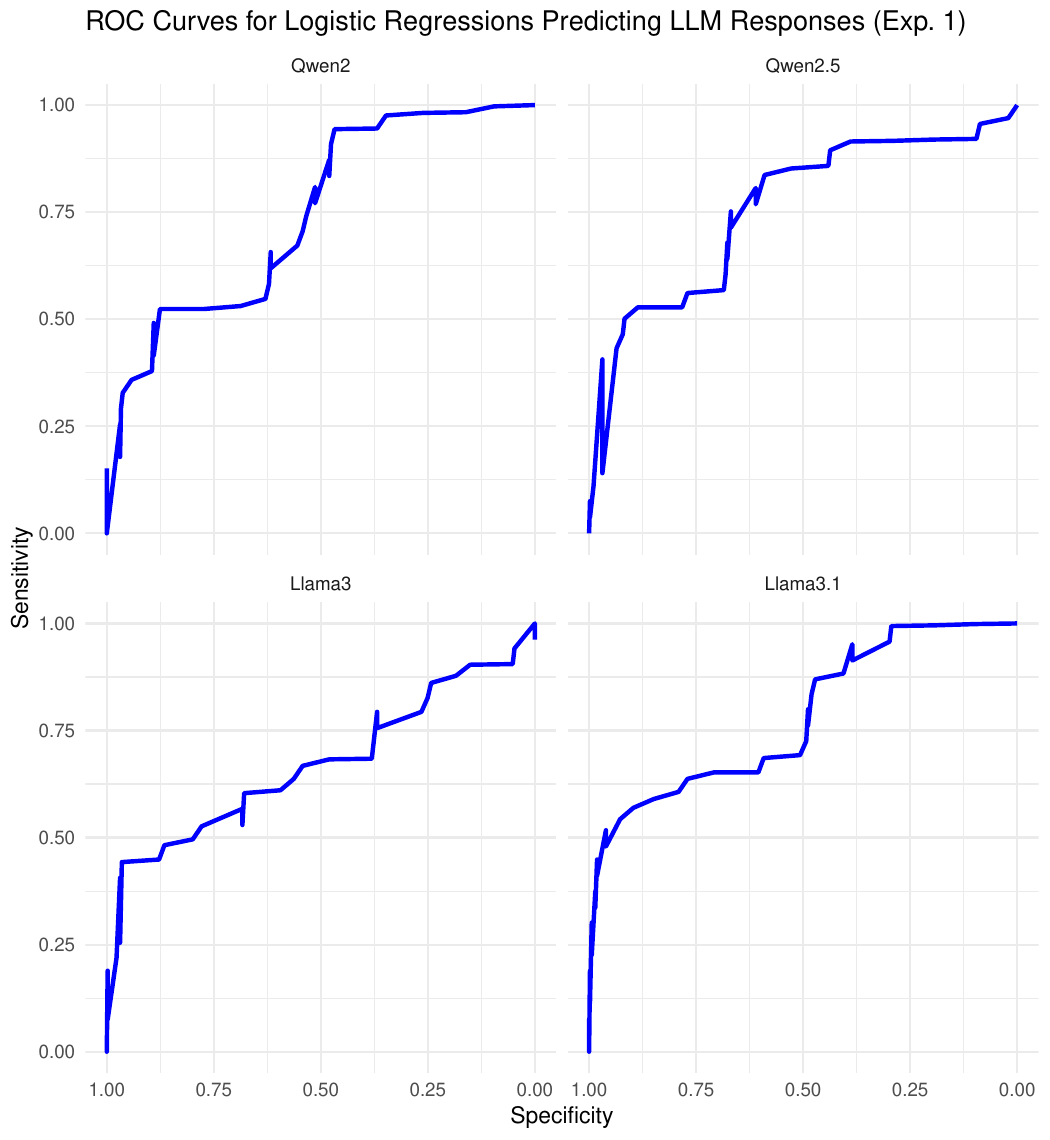}
\caption{Receiver operating characteristic (ROC) curves for logistic regression models predicting LLM response data from Experiment~1.}
\label{fig:exp1-roc}
\end{figure}

All factors in the logistic regression models were contrast-coded using sum-to-zero effect coding. The precise coding scheme applied for each factor is reported in Table \ref{table: exp1-cod}, in \ref{app:contrast:exp1}. The regression results are summarized in Table~\ref{table: exp1-Qwen2} (Qwen2), Table~\ref{table: exp1-Qwen2.5} (Qwen2.5), Table~\ref{table: exp1-Llama3} (Llama3), and Table~\ref{table: exp1-Llama3.1} (Llama3.1), respectively. Receiver operating characteristic (ROC) curves for the optimally fitting logistic regression models for each LLM class are shown in Figure~\ref{fig:exp1-roc}, Appendix~\ref{app:roccurves:exp2}

\subsection{Experiment 2}
\label{app:logistic:exp2}

In the logistic regression analyses for Experiment~2, the factors number of parameters, epistemic certainty, statement type, and QA format are treated as main effects. They are coded using sum-to-zero effect coding (see details on contrast coding in Table \ref{table: exp2-cod}, Appendix \ref{app‌:contrast:exp2}). The two-way interactions between number of parameters and any of the remaining factors are also assessed.

A separate logistic regression model is fitted for each examined class of LLMs (Qwen2, Qwen2.5, Llama3, Llama3.1). Using the \emph{buildmer} package (\citealp{Voeten2024}) for R, we conduct an AIC-based model selection process. Its main purpose is to check whether it is necessary to include any random effects which would be theoretically justified by the design. We consider by-item random intercepts as well as by-item random slopes for the factor number of parameters. As the model selection analyses reveal, however, in all cases the retained optimal model does not include any random effects, i.e., is just a simple logistic regression model.

\begin{figure}[ht!]
\centering
\includegraphics[width=0.9\textwidth]{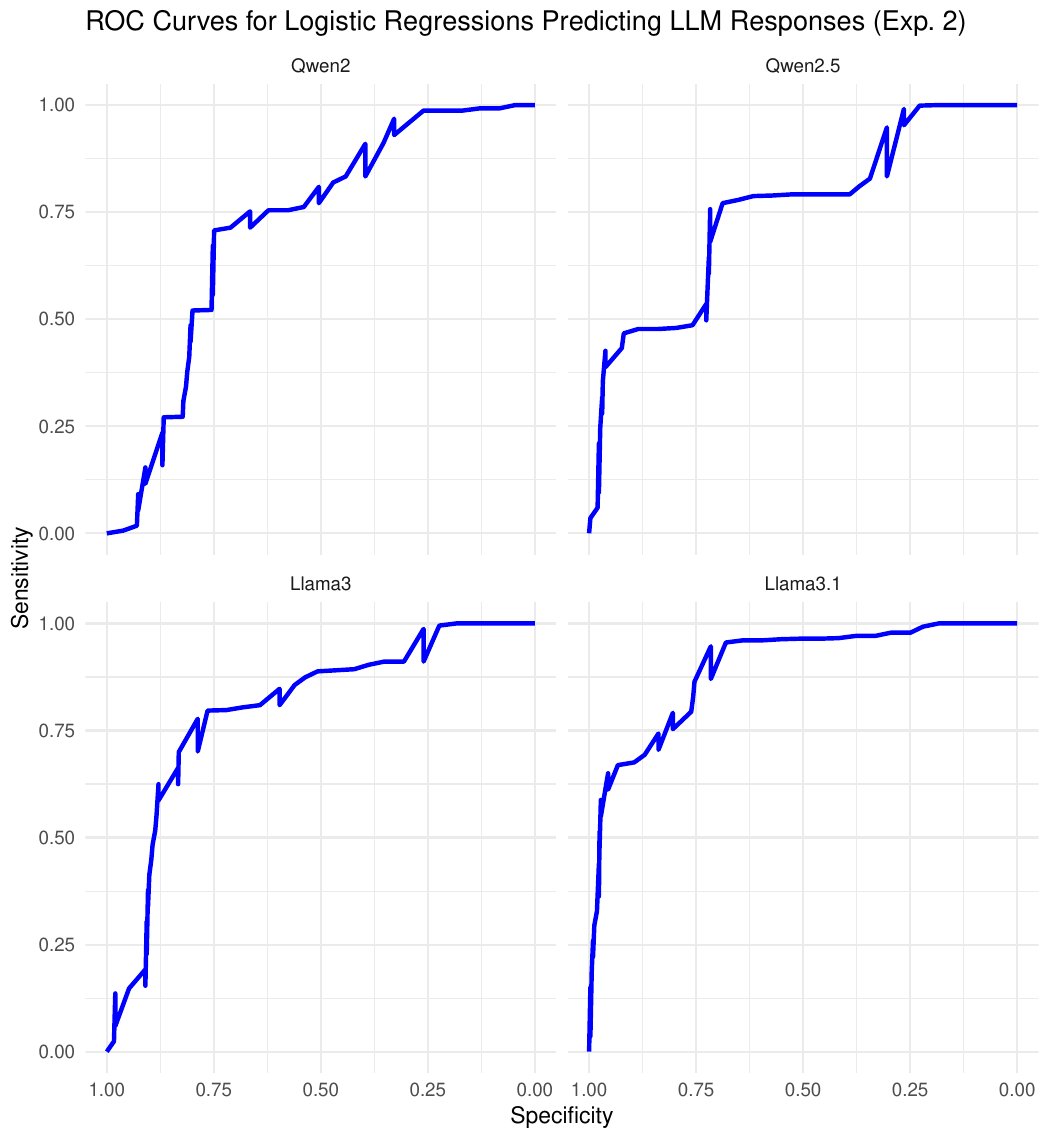}
\caption{Receiver operating characteristic (ROC) curves for logistic regression models predicting LLM response data from Experiment~2.}
\label{fig:exp2-roc}
\end{figure}

In Appendix \ref{app:statistical:exp2}, all results of the logistic regression analyses for Experiment 2 are reported; see Tables \ref{table: exp2-Qwen2} (Qwen2), \ref{table: exp2-Qwen2.5} (Qwen2.5), \ref{table: exp2-Llama3} (Llama3), and \ref{table: exp2-Llama3.1} (Llama3.1), respectively. ROC curves for the optimally fitting logistic regression models for each LLM class are shown in Figure~\ref{fig:exp2-roc}, Appendix~\ref{app:roccurves:exp2}.

\clearpage
\section{Further Plots of Response Accuracy by Condition}
\label{app:figures}

\subsection{Experiment 1}
\label{app‌:figures:exp1}

\begin{figure}[ht!]
\centering
\includegraphics[width=0.48\textwidth]{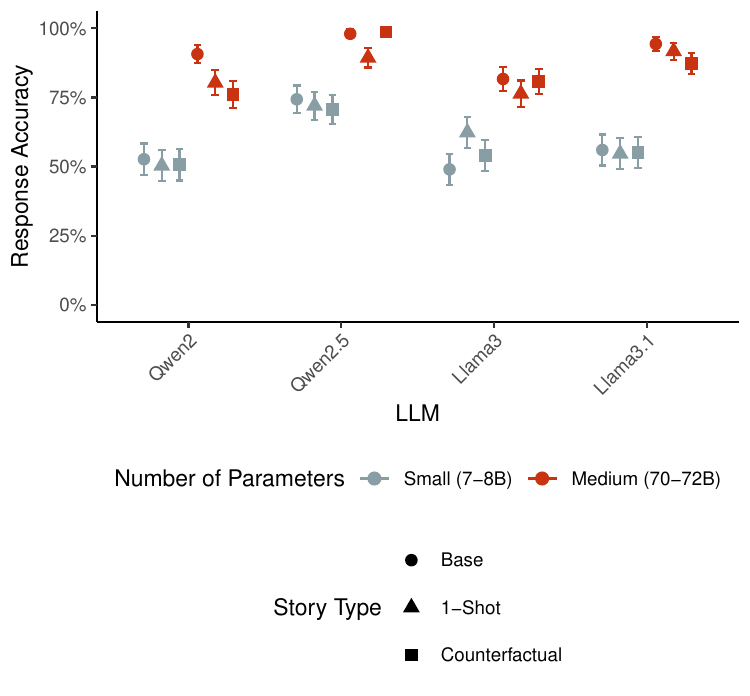}
\caption{Response accuracy in Experiment 1 by LLM, number of parameters, and story type. Error bars represent 95\% confidence intervals.}
\label{fig:exp1-acc-dotplot-exp}
\end{figure}

\begin{figure}[ht!]
\centering
\includegraphics[width=0.48\textwidth]{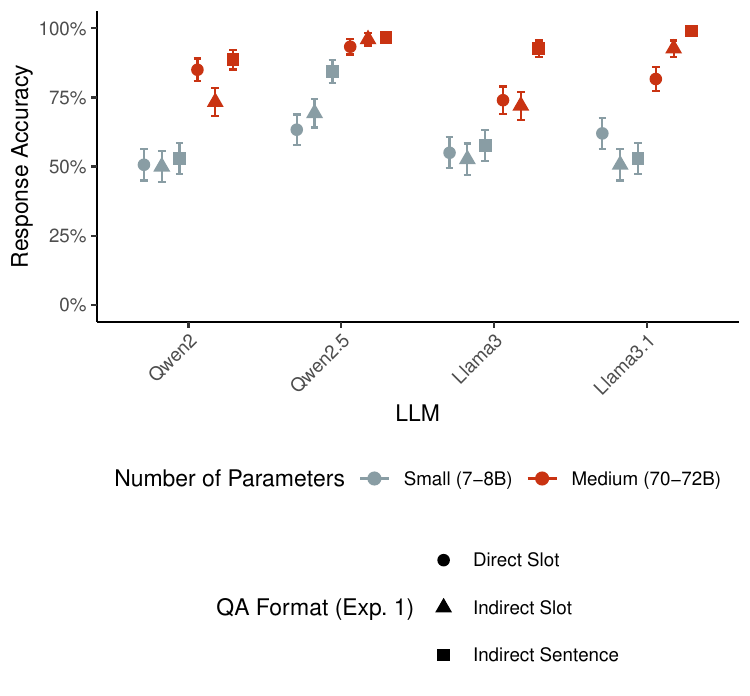}
\caption{Response accuracy in Experiment 1 by LLM, number of parameters, and QA format. Error bars represent 95\% confidence intervals.}
\label{fig:exp1-acc-dotplot-qa}
\end{figure}

%\FloatBarrier

%\clearpage
\subsection{Experiment 2}
\label{app‌:figures:exp2}

\begin{figure}[ht!]
\centering
\includegraphics[width=0.48\textwidth]{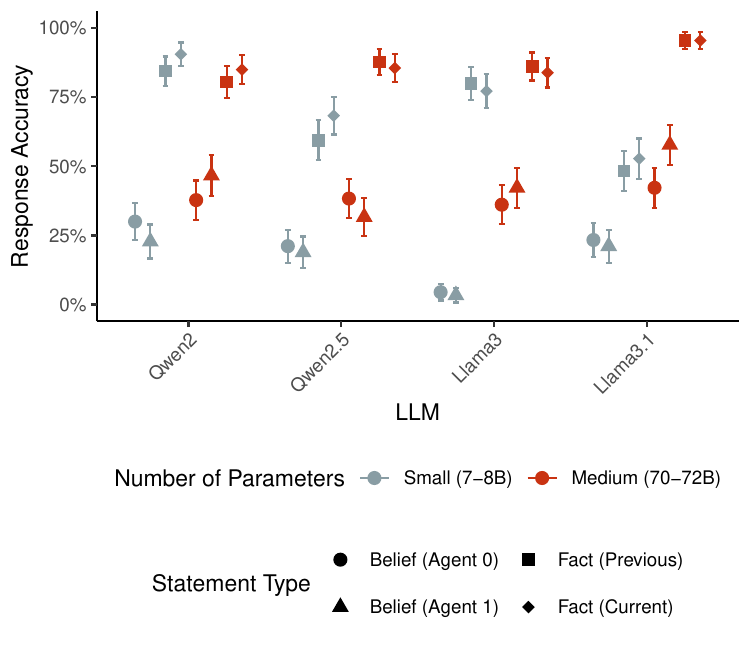}
\caption{Response accuracy in Experiment 2 by LLM, number of parameters, and statement type. Error bars represent 95\% confidence intervals.}
\label{fig:exp2-acc-dotplot-statement}
\end{figure}

%\begin{figure*}[ht!]
%\centering
%\includegraphics[width=\textwidth]{figures/exp2_acc_dotplot_n_pars_belief.pdf}
%\caption{Response accuracy by LLM, number of parameters, and belief levels. Error bars represent 95\% confidence intervals.}
%\label{fig:exp2-acc-dotplot-belief}
%\end{figure*}

\begin{figure}[ht!]
\centering
\includegraphics[width=0.48\textwidth]{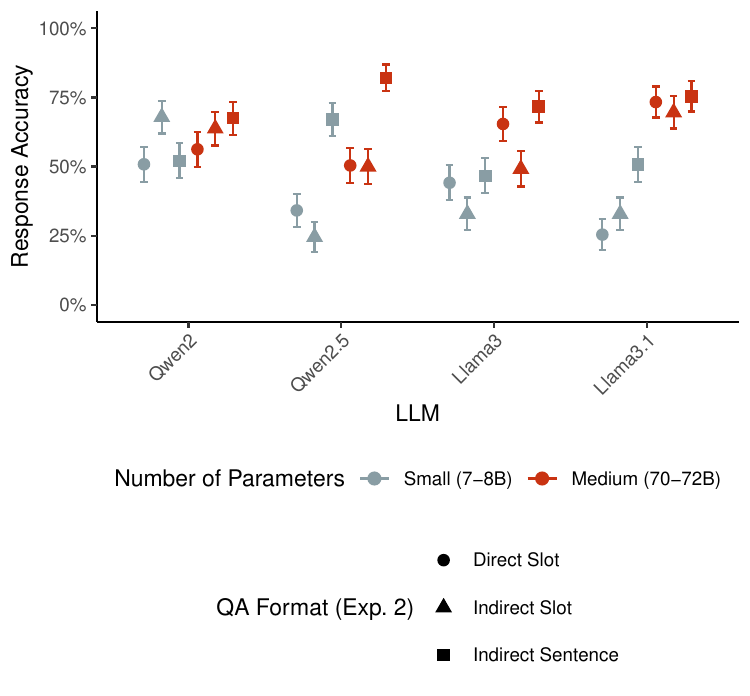}
\caption{Response accuracy in Experiment 2 by LLM, number of parameters, and modal condition. Error bars represent 95\% confidence intervals.}
\label{fig:exp2-acc-dotplot-qa}
\end{figure}

%\FloatBarrier

%~
%\clearpage
%\twocolumn

\clearpage
\onecolumn
\section{Story Templates and Examples}
\label{app:template}
% examples

\subsection{Experiment 1}
\label{app:template:exp1}

There are five templates to generate stories in Experiment~1: \textit{the hidden object}, \textit{Whodunnit}, \textit{city travel}, \textit{the grocery store's promotion}, and \textit{fictional characters}. \textit{The hidden object} is a classic task from \citet{hirst1982acquisition}, \citet{ noveck1996children}, \citet{ ozturk2015acquisition}. The textual template of \textit{Whodunnit} is inspired by \citet{wu2024whodunnit} and \citet{jin2024marple}. The others are designed by ourselves.

%\begin{landscape}
%\clearpage

\small
\begin{longtable}{p{0.08\linewidth}|p{0.07\linewidth}|p{0.07\linewidth}|p{0.6\linewidth}|p{0.05\linewidth}}

\hline
\toprule
\textbf{Types} & \textbf{N/P condition} & \textbf{QA format} & \textbf{Prompt} & \textbf{Answer} \\ \hline
\midrule
base & N & direct slot & There are three boxes in a room: a black box, a purple box and a blue box. A peanut is hidden in one of these boxes. It is known that the peanut is not in the black box and not in the purple box. The peanut \_ be in the blue box. Which word or expression is more adequate to describe the situation after filling in the slot:  may or has to? Please ONLY respond with may or has to as your answer. & has to \\ \hline
base & P & direct slot & There are three boxes in a room: a black box, a purple box and a blue box. A peanut is hidden in one of these boxes. It is known that the peanut is not in the black box. The peanut \_ be in the blue box. Which word or expression is more adequate to describe the situation after filling in the slot:  may or has to? Please ONLY respond with may or has to as your answer. & may \\ \hline
base & N & indirect slot & There are three boxes in a room: a black box, a purple box and a blue box. A peanut is hidden in one of these boxes. It is known that the peanut is not in the black box and not in the purple box. The peanut \_ be in the blue box. Which word or expression is more adequate to describe the situation after filling in the slot:  1 may or 2 has to? Please ONLY respond with 1 or 2 as your answer. & 2 \\ \hline
base & P & indirect slot & There are three boxes in a room: a black box, a purple box and a blue box. A peanut is hidden in one of these boxes. It is known that the peanut is not in the black box. The peanut \_ be in the blue box. Which word or expression is more adequate to describe the situation after filling in the slot:  1 may or 2 has to? Please ONLY respond with 1 or 2 as your answer. & 1 \\ \hline
base & N & indirect sentence & There are three boxes in a room: a black box, a purple box and a blue box. A peanut is hidden in one of these boxes. It is known that the peanut is not in the black box and not in the purple box. Which sentence is more adequate to describe the situation? 1 The peanut may be in the blue box. 2 The peanut has to be in the blue box. Please ONLY respond with 1 or 2 as your answer. & 2 \\ \hline
base & P & indirect sentence & There are three boxes in a room: a black box, a purple box and a blue box. A peanut is hidden in one of these boxes. It is known that the peanut is not in the black box. Which sentence is more adequate to describe the situation? 1 The peanut may be in the blue box. 2 The peanut has to be in the blue box. Please ONLY respond with 1 or 2 as your answer. & 1 \\ \hline
1-shot & N & direct slot & There are three boxes in a room: a yellow box, a red box and a gray box. A peanut is hidden in one of these boxes. If the peanut is not in the gray box, the peanut may be in the yellow box or red box. If the peanut is not in the gray and the red box, the peanut has to be in the yellow box. Now there is another case. There are three boxes in a room: a black box, a purple box and a blue box. A peanut is hidden in one of these boxes. It is known that the peanut is not in the black box and not in the purple box. The peanut \_ be in the blue box. Which word or expression is more adequate to describe the situation after filling in the slot:  may or has to? Please ONLY respond with may or has to as your answer. & has to \\ \hline
1-shot & P & direct slot & There are three boxes in a room: a yellow box, a red box and a gray box. A peanut is hidden in one of these boxes. If the peanut is not in the gray box, the peanut may be in the yellow box or red box. If the peanut is not in the gray and the red box, the peanut has to be in the yellow box. Now there is another case. There are three boxes in a room: a black box, a purple box and a blue box. A peanut is hidden in one of these boxes. It is known that the peanut is not in the black box. The peanut \_ be in the blue box. Which word or expression is more adequate to describe the situation after filling in the slot: may or has to? Please ONLY respond with may or has to as your answer. & may \\ \hline
1-shot & N & indirect slot & There are three boxes in a room: a yellow box, a red box and a gray box. A peanut is hidden in one of these boxes. If the peanut is not in the gray box, the peanut may be in the yellow box or red box. If the peanut is not in the gray and the red box, the peanut has to be in the yellow box. Now there is another case. There are three boxes in a room: a black box, a purple box and a blue box. A peanut is hidden in one of these boxes. It is known that the peanut is not in the black box and not in the purple box. The peanut \_ be in the blue box. Which word or expression is more adequate to describe the situation after filling in the slot:  1 may or 2 has to? Please ONLY respond with 1 or 2 as your answer. & 2 \\ \hline
1-shot & P & indirect slot & There are three boxes in a room: a yellow box, a red box and a gray box. A peanut is hidden in one of these boxes. If the peanut is not in the gray box, the peanut may be in the yellow box or red box. If the peanut is not in the gray and the red box, the peanut has to be in the yellow box. Now there is another case. There are three boxes in a room: a black box, a purple box and a blue box. A peanut is hidden in one of these boxes. It is known that the peanut is not in the black box. The peanut \_ be in the blue box. Which word or expression is more adequate to describe the situation after filling in the slot:  1 may or 2 has to? Please ONLY respond with 1 or 2 as your answer. & 1 \\ \hline
1-shot & N & indirect sentence & There are three boxes in a room: a yellow box, a red box and a gray box. A peanut is hidden in one of these boxes. If the peanut is not in the gray box, the peanut may be in the yellow box or red box. If the peanut is not in the gray and the red box, the peanut has to be in the yellow box. Now there is another case. There are three boxes in a room: a black box, a purple box and a blue box. A peanut is hidden in one of these boxes. It is known that the peanut is not in the black box and not in the purple box. Which sentence is more adequate to describe the situation? 1 The peanut may be in the blue box. 2 The peanut has to be in the blue box. Please ONLY respond with 1 or 2 as your answer. & 2 \\ \hline
1-shot & P & indirect sentence & There are three boxes in a room: a yellow box, a red box and a gray box. A peanut is hidden in one of these boxes. If the peanut is not in the gray box, the peanut may be in the yellow box or red box. If the peanut is not in the gray and the red box, the peanut has to be in the yellow box. Now there is another case. There are three boxes in a room: a black box, a purple box and a blue box. A peanut is hidden in one of these boxes. It is known that the peanut is not in the black box. Which sentence is more adequate to describe the situation? 1 The peanut may be in the blue box. 2 The peanut has to be in the blue box. Please ONLY respond with 1 or 2 as your answer. & 1 \\ \hline
counter-factual & N & direct slot & There are three boxes in a room: a black box, a purple box and a blue box. A peanut is hiding in the black box, because yesterday Tom moved the peanut from the purple box to the black box. If he hadn't moved the peanut yesterday, imagine which box the peanut would be in today. The peanut \_ be in the purple box. Which word or expression is more adequate to describe the situation after filling in the slot: may or has to? Please ONLY respond with may or has to as your answer. & has to \\ \hline
counter-factual & P & direct slot & There are three boxes in a room: a black box, a purple box and a blue box. A peanut is hiding in the black box, because yesterday Tom moved the peanut from one of the other two boxes to the black box. If he hadn't moved the peanut yesterday, imagine which box the peanut would be in today. The peanut \_ be in the purple box. Which word or expression is more adequate to describe the situation after filling in the slot: may or has to? Please ONLY respond with may or has to as your answer. & may \\ \hline
counter-factual & N & indirect slot & There are three boxes in a room: a black box, a purple box and a blue box. A peanut is hiding in the black box, because yesterday Tom moved the peanut from the purple box to the black box. If he hadn't moved the peanut yesterday, imagine which box the peanut would be in today. The peanut \_ be in the purple box. Which word or expression is more adequate to describe the situation after filling in the slot: 1 may or 2 has to? Please ONLY respond with 1 or 2 as your answer. & 2 \\ \hline
counter-factual & P & indirect slot & There are three boxes in a room: a black box, a purple box and a blue box. A peanut is hiding in the black box, because yesterday Tom moved the peanut from one of the other two boxes to the black box. If he hadn't moved the peanut yesterday, imagine which box the peanut would be in today. The peanut \_ be in the purple box. Which word or expression is more adequate to describe the situation after filling in the slot: 1 may or 2 has to? Please ONLY respond with 1 or 2 as your answer. & 1 \\ \hline
counter-factual & N & indirect sentence & There are three boxes in a room: a black box, a purple box and a blue box. A peanut is hiding in the black box, because yesterday Tom moved the peanut from the purple box to the black box. If he hadn't moved the peanut yesterday, imagine which box the peanut would be in today. Which sentence is more adequate to describe the imagined situation? 1 The peanut may be in the purple box. 2 The peanut has to be in the purple box. Please ONLY respond with 1 or 2 as your answer. & 2 \\ \hline
counter-factual & P & indirect sentence & There are three boxes in a room: a black box, a purple box and a blue box. A peanut is hiding in the black box, because yesterday Tom moved the peanut from one of the other two boxes to the black box. If he hadn't moved the peanut yesterday, imagine which box the peanut would be in today. Which sentence is more adequate to describe the imagined situation? 1 The peanut may be in the blue box. 2 The peanut has to be in the blue box. Please ONLY respond with 1 or 2 as your answer. & 1 \\ \hline
\end{longtable}
%\end{landscape}

\subsection{Experiment 2}
\label{app‌:template:exp2}
A ToM story with eight statements in direct slot format is demonstrated below.

\small
\begin{longtable}{p{0.15\linewidth}|p{0.1\linewidth}|p{0.52\linewidth}|p{0.08\linewidth}}
%\begin{tabular}
\hline
\toprule
 \textbf{Statements} & \textbf{Certainty} &  \textbf{Prompt} & \textbf{Answer} \\ \hline
\midrule

Fact (previous) & Not low & You witness a sequence of events: 1 Hannah entered the hall. 2 Amelia entered the hall. 3 The corn is in the drawer. 4 The drawer is in the hall. 5 Amelia hates the grapes. 6 Amelia hates the stockings. 7 Hannah exited the hall. 8 Amelia moved the corn to the bathtub. 9 The bathtub is in the hall. 10 Hannah entered the hall. 11 Charlotte entered the hall.  You use the first-person perspective to report the situation: \colorbox{lime}{I \_ the corn was in the drawer}. Which word or expression is more adequate to describe the situation after filling in the slot: know, believe or doubt? Please ONLY respond with know, believe or doubt as your answer. & know \\ \hline
Fact (previous) & Low & You witness a sequence of events: 1 Hannah entered the hall. 2 Amelia entered the hall. 3 The corn is in the drawer. 4 The drawer is in the hall. 5 Amelia hates the grapes. 6 Amelia hates the stockings. 7 Hannah exited the hall. 8 Amelia moved the corn to the bathtub. 9 The bathtub is in the hall. 10 Hannah entered the hall. 11 Charlotte entered the hall.  You use the first-person perspective to report the situation: \colorbox{pink}{I \_ the corn was in the box}. Which word or expression is more adequate to describe the situation after filling in the slot: know, believe or doubt? Please ONLY respond with know, believe or doubt as your answer. & doubt \\ \hline
Fact (current) & Not low & You witness a sequence of events: 1 Hannah entered the hall. 2 Amelia entered the hall. 3 The corn is in the drawer. 4 The drawer is in the hall. 5 Amelia hates the grapes. 6 Amelia hates the stockings. 7 Hannah exited the hall. 8 Amelia moved the corn to the bathtub. 9 The bathtub is in the hall. 10 Hannah entered the hall. 11 Charlotte entered the hall.  You use the first-person perspective to report the situation: \colorbox{lime}{I \_ the corn is in the bathtub}. Which word or expression is more adequate to describe the situation after filling in the slot: know, believe or doubt? Please ONLY respond with know, believe or doubt as your answer. & know \\ \hline
Fact (current) & Low  & You witness a sequence of events: 1 Hannah entered the hall. 2 Amelia entered the hall. 3 The corn is in the drawer. 4 The drawer is in the hall. 5 Amelia hates the grapes. 6 Amelia hates the stockings. 7 Hannah exited the hall. 8 Amelia moved the corn to the bathtub. 9 The bathtub is in the hall. 10 Hannah entered the hall. 11 Charlotte entered the hall.  You use the first-person perspective to report the situation: \colorbox{pink}{I \_ the corn is in the box}. Which word or expression is more adequate to describe the situation after filling in the slot: know, believe or doubt? Please ONLY respond with know, believe or doubt as your answer. & doubt \\ \hline 
Belief (agent 0) & Not low & You witness a sequence of events: 1 Hannah entered the hall. 2 Amelia entered the hall. 3 The corn is in the drawer. 4 The drawer is in the hall. 5 Amelia hates the grapes. 6 Amelia hates the stockings. 7 Hannah exited the hall. 8 Amelia moved the corn to the bathtub. 9 The bathtub is in the hall. 10 Hannah entered the hall. 11 Charlotte entered the hall.  You use the first-person perspective to report the situation: \colorbox{cyan}{I \_ Amelia will look for the corn in the bathtub}. Which word or expression is more adequate to describe the situation after filling in the slot: know, believe or doubt? Please ONLY respond with know, believe or doubt as your answer. & believe \\ \hline 
Belief (agent 0) & Low & You witness a sequence of events: 1 Hannah entered the hall. 2 Amelia entered the hall. 3 The corn is in the drawer. 4 The drawer is in the hall. 5 Amelia hates the grapes. 6 Amelia hates the stockings. 7 Hannah exited the hall. 8 Amelia moved the corn to the bathtub. 9 The bathtub is in the hall. 10 Hannah entered the hall. 11 Charlotte entered the hall.  You use the first-person perspective to report the situation: \colorbox{orange}{I \_ Amelia will look for the corn in the drawer}. Which word or expression is more adequate to describe the situation after filling in the slot: know, believe or doubt? Please ONLY respond with know, believe or doubt as your answer. & doubt \\ \hline
Belief (agent 1) & Not low & You witness a sequence of events: 1 Hannah entered the hall. 2 Amelia entered the hall. 3 The corn is in the drawer. 4 The drawer is in the hall. 5 Amelia hates the grapes. 6 Amelia hates the stockings. 7 Hannah exited the hall. 8 Amelia moved the corn to the bathtub. 9 The bathtub is in the hall. 10 Hannah entered the hall. 11 Charlotte entered the hall.  You use the first-person perspective to report the situation: \colorbox{cyan}{I \_ Hannah will look for the corn in the drawer}. Which word or expression is more adequate to describe the situation after filling in the slot: know, believe or doubt? Please ONLY respond with know, believe or doubt as your answer. & believe \\ \hline
Belief (agent 1) & Low & You witness a sequence of events: 1 Hannah entered the hall. 2 Amelia entered the hall. 3 The corn is in the drawer. 4 The drawer is in the hall. 5 Amelia hates the grapes. 6 Amelia hates the stockings. 7 Hannah exited the hall. 8 Amelia moved the corn to the bathtub. 9 The bathtub is in the hall. 10 Hannah entered the hall. 11 Charlotte entered the hall.  You use the first-person perspective to report the situation: \colorbox{orange}{I \_ Hannah will look for the corn in the bathtub}. Which word or expression is more adequate to describe the situation after filling in the slot: know, believe or doubt? Please ONLY respond with know, believe or doubt as your answer. & doubt \\ \hline
\end{longtable}

\clearpage
\onecolumn
\section{Statistical Coefficients From Logistic Regression Analyses}
\label{app:statistical}

\subsection{Experiment 1}
\label{app:statistical:exp1}

\begin{table*}[h]
\small
\centering
\begin{tabular}{llrcrr}
\toprule
& & \multicolumn{4}{c}{\textbf{Correct Response}} \\
\multicolumn{2}{l}{Predictor} & $b~~~$ & $SE$ & $z~~~$ & $p~~~~~$ \\
\midrule
\multicolumn{2}{l}{(Intercept)} & 0.87 & 0.06 & 14.37 & <\,.001 \\
\multicolumn{2}{l}{Number of Parameters [Medium\,>\,Small]} & 1.63 & 0.12 & 13.46 & <\,.001 \\
\multicolumn{2}{l}{Modal Condition [Necessity\,>\,Possibility]} & 0.87 & 0.12 & 7.59 & <\,.001 \\
\multicolumn{2}{l}{Story Type 1-Shot [>\,Base]} & $-$0.21 & 0.16 & $-$1.31 & <\,.001 \\
\multicolumn{2}{l}{Story Type Counterfactual [>\,Base]} & $-$0.50 & 0.16 & $-$3.15 & .002 \\
\multicolumn{2}{l}{QA Format Indirect Slot [>\,Direct Slot]} & $-$0.66 & 0.16 & $-$4.23 & <\,.001 \\
\multicolumn{2}{l}{QA Format Indirect Sentence [>\,Direct Slot]} & 0.55 & 0.17 & 3.18 & .002 \\
\multicolumn{2}{l}{Number of Parameters $\times$ Modal Condition} & $-$0.93 & 0.23 & $-$4.05 & <\,.001 \\
\multicolumn{2}{l}{Number of Parameters $\times$ Story Type Counterfactual} & $-$1.03 & 0.29 & $-$3.52 & <\,.001 \\
\multicolumn{2}{l}{Number of Parameters $\times$ QA Format Indirect Slot} & $-$1.11 & 0.31 & $-$3.53 & <\,.001 \\
\multicolumn{2}{l}{Number of Parameters $\times$ QA Format Indirect Sentence} & 0.78 & 0.35 & 2.26 & .024 \\
\midrule
Observations & 1800 & & & \\
$R^2_{\mathrm{\,Tjur}}$ & .186 & & & \\
AIC & 1956.6 & & & \\
\bottomrule
\end{tabular}
\caption{Logistic regression results on \textbf{Qwen2-7B/72B} data from Experiment 1, derived from the optimal regression model selected by AIC.}
\label{table: exp1-Qwen2}
\end{table*}

\begin{table*}[h]
\small
\centering
\begin{tabular}{llrcrr}
\toprule
& & \multicolumn{4}{c}{\textbf{Correct Response}} \\
\multicolumn{2}{l}{Predictor} & $b~~~$ & $SE$ & $z~~~$ & $p~~~~~$ \\
\midrule
\multicolumn{2}{l}{(Intercept)} & 2.67 & 0.18 & 14.86 & <\,.001 \\
\multicolumn{2}{l}{Number of Parameters [Medium\,>\,Small]} & 3.15 & 0.36 & 8.84 & <\,.001 \\
\multicolumn{2}{l}{Modal Condition [Necessity\,>\,Possibility]} & 1.99 & 0.32 & 6.31 & <\,.001 \\
\multicolumn{2}{l}{Story Type 1-Shot [>\,Base]} & $-$1.41 & 0.28 & $-$5.05 & <\,.001 \\
\multicolumn{2}{l}{Story Type Counterfactual [>\,Base]} & 0.81 & 0.39 & 2.09 & .036 \\
\multicolumn{2}{l}{QA Format Indirect Slot [>\,Direct Slot]} & $-$0.31 & 0.20 & $-$1.52 & .128 \\
\multicolumn{2}{l}{QA Format Indirect Sentence [>\,Direct Slot]} & 1.30 & 0.23 & 5.78 & <\,.001 \\
\multicolumn{2}{l}{Number of Parameters $\times$ Modal Condition} & 1.54 & 0.63 & 2.44 & .015 \\
\multicolumn{2}{l}{Number of Parameters $\times$ Story Type 1-Shot} & $-$2.74 & 0.56 & $-$4.90 & <\,.001 \\
\multicolumn{2}{l}{Number of Parameters $\times$ Story Type Counterfactual} & 1.99 & 0.78 & 2.57 & .010 \\
\midrule
Observations & 1800 & & & \\
$R^2_{\mathrm{\,Tjur}}$ & .206 & & & \\
AIC & 1246.0 & & & \\
\bottomrule
\end{tabular}
\caption{Logistic regression results on \textbf{Qwen2.5-7B/72B} data from Experiment 1, derived from the optimal regression model selected by AIC.}
\label{table: exp1-Qwen2.5}
\end{table*}

\begin{table*}[h]
\small
\centering
\begin{tabular}{llrcrr}
\toprule
& & \multicolumn{4}{c}{\textbf{Correct Response}} \\
\multicolumn{2}{l}{Predictor} & $b~~~$ & $SE$ & $z~~~$ & $p~~~~~$ \\
\midrule
\multicolumn{2}{l}{(Intercept)} & 1.30 & 0.10 & 13.20 & <\,.001 \\
\multicolumn{2}{l}{Number of Parameters [Medium\,>\,Small]} & 2.17 & 0.20 & 11.08 & <\,.001 \\
\multicolumn{2}{l}{Modal Condition [Necessity\,>\,Possibility]} & 1.79 & 0.19 & 9.44 & <\,.001 \\
\multicolumn{2}{l}{Story Type 1-Shot [>\,Base]} & 0.04 & 0.17 & 0.23 & .822 \\
\multicolumn{2}{l}{Story Type Counterfactual [>\,Base]} & $-$0.01 & 0.16 & $-$0.04 & .971 \\
\multicolumn{2}{l}{QA Format Indirect Slot [>\,Direct Slot]} & $-$0.79 & 0.17 & $-$4.69 & <\,.001 \\
\multicolumn{2}{l}{QA Format Indirect Sentence [>\,Direct Slot]} & 1.34 & 0.20 & 6.86 & <\,.001 \\
\multicolumn{2}{l}{Number of Parameters $\times$ Modal Condition} & 3.66 & 0.38 & 9.63 & <\,.001 \\
\multicolumn{2}{l}{Number of Parameters $\times$ Story Type 1-Shot} & $-$1.02 & 0.30 & $-$3.44 & <\,.001 \\
\multicolumn{2}{l}{Number of Parameters $\times$ QA Format Indirect Slot} & $-$1.19 & 0.34 & $-$3.51 & <\,.001 \\
\multicolumn{2}{l}{Number of Parameters $\times$ QA Format Indirect Sentence} & 2.27 & 0.39 & 5.79 & <\,.001 \\
\midrule
Observations & 1800 & & & \\
$R^2_{\mathrm{\,Tjur}}$ & .189 & & & \\
AIC & 1866.7 & & & \\
\bottomrule
\end{tabular}
\caption{ Logistic regression results on \textbf{Llama3-8B/70B} data from Experiment 1, derived from the optimal regression model selected by AIC.}
\label{table: exp1-Llama3}
\end{table*}

\begin{table*}[h]
\small
\centering
\begin{tabular}{llrcrr}
\toprule
& & \multicolumn{4}{c}{\textbf{Correct Response}} \\
\multicolumn{2}{l}{Predictor}& $b~~~$ & $SE$ & $z~~~$ & $p~~~~~$ \\
\midrule
\multicolumn{2}{l}{(Intercept)} & 1.70 & 0.11 & 15.40 & <\,.001 \\
\multicolumn{2}{l}{Number of Parameters [Medium\,>\,Small]} & 2.96 & 0.22 & 13.45 & <\,.001 \\
\multicolumn{2}{l}{Modal Condition [Necessity\,>\,Possibility]} & 1.14 & 0.17 & 6.82 & <\,.001 \\
\multicolumn{2}{l}{Story Type 1-Shot [>\,Base]} & $-$0.02 & 0.17 & $-$0.14 & .887 \\
\multicolumn{2}{l}{Story Type Counterfactual [>\,Base]} & $-$0.52 & 0.19 & $-$2.67 & .008 \\
\multicolumn{2}{l}{QA Format Indirect Slot [>\,Direct Slot]} & $-$0.41 & 0.18 & $-$2.33 & .020 \\
\multicolumn{2}{l}{QA Format Indirect Sentence [>\,Direct Slot]} & 1.54 & 0.28 & 5.61 & <\,.001 \\
\multicolumn{2}{l}{Number of Parameters $\times$ Modal Condition} & 1.16 & 0.33 & 3.48 & <\,.001 \\
\multicolumn{2}{l}{Number of Parameters $\times$ Story Type Counterfactual} & $-$0.98 & 0.36 & $-$2.70 & .007 \\
\multicolumn{2}{l}{Number of Parameters $\times$ QA Format Indirect Sentence} & 3.43 & 0.48 & 7.19 & <\,.001 \\
\midrule
Observations & 1800 & & & \\
$R^2_{\mathrm{\,Tjur}}$ & .210 & & & \\
AIC & 1659.4 & & & \\
\bottomrule
\end{tabular}
\caption{Logistic regression results on \textbf{Llama3.1-8B/70B} data from Experiment 1, derived from the optimal regression model selected by AIC.}
\label{table: exp1-Llama3.1}
\end{table*}

~
%\clearpage
\subsection{Experiment 2}
\label{app:statistical:exp2}

\begin{table*}[h]
\small
\centering
\begin{tabular}{llrcrr}
\toprule
& & \multicolumn{4}{c}{\textbf{Correct Response}} \\
\multicolumn{2}{l}{Predictor} & $b~~~$ & $SE$ & $z~~~$ & $p~~~~~$ \\
\midrule
\multicolumn{2}{l}{(Intercept)} & 0.62 & 0.07 & 8.56 & <\,.001 \\
\multicolumn{2}{l}{Number of Parameters [Medium\,>\,Small]} & 0.08 & 0.15 & 0.52 & .604 \\
\multicolumn{2}{l}{Epistemic Certainty [Not Low\,>\,Low]} & 1.64 & 0.16 & 9.98 & <\,.001 \\
\multicolumn{2}{l}{Statement Type Fact [>\,Belief]} & 3.02 & 0.17 & 17.34 & <\,.001 \\
\multicolumn{2}{l}{Statement Type Agent 1 [>\,Agent 0]} & $-$0.05 & 0.18 & $-$0.31 & .757 \\
\multicolumn{2}{l}{Statement Type Current [>\,Previous]} & 0.49 & 0.23 & 2.16 & .031 \\
\multicolumn{2}{l}{QA Format Indirect Slot [>\,Direct Slot]} & 0.99 & 0.21 & 4.82 & <\,.001 \\
\multicolumn{2}{l}{QA Format Indirect Sentence [>\,Direct Slot]} & $-$0.14 & 0.20 & $-$0.70 & .481 \\
\multicolumn{2}{l}{Number of Parameters $\times$ Epistemic Certainty} & $-$1.61 & 0.33 & $-$4.89 & <\,.001 \\
\multicolumn{2}{l}{Number of Parameters $\times$ Statement Type Fact} & $-$2.06 & 0.35 & $-$5.92 & <\,.001 \\
\multicolumn{2}{l}{Number of Parameters $\times$ Statement Type Agent 1} & 0.88 & 0.35 & 2.49 & .013 \\
\multicolumn{2}{l}{Number of Parameters $\times$ Statement Type Current} & $-$0.33 & 0.45 & $-$0.73 & .466 \\
\multicolumn{2}{l}{Number of Parameters $\times$ QA Format Indirect Slot} & $-$1.72 & 0.41 & $-$4.19 & <\,.001 \\
\multicolumn{2}{l}{Number of Parameters $\times$ QA Format Indirect Sentence} & 1.37 & 0.39 & 3.47 & <\,.001 \\
\midrule
Observations & 1440 & & & \\
$R^2_{\mathrm{\,Tjur}}$ & .384 & & & \\
AIC & 1354.3 & & & \\
\bottomrule
\end{tabular}
\caption{Logistic regression results on \textbf{Qwen2-7B/72B} data from Experiment 2, derived from the optimal regression model selected by AIC.}
\label{table: exp2-Qwen2}
\end{table*}

\begin{table*}[h]
\small
\centering
\begin{tabular}{llrcrr}
\toprule
& & \multicolumn{4}{c}{\textbf{Correct Response}} \\
\multicolumn{2}{l}{Predictor} & $b~~~$ & $SE$ & $z~~~$ & $p~~~~~$ \\
\midrule
\multicolumn{2}{l}{(Intercept)} & $-$0.01 & 0.09 & $-$0.10 & .921 \\
\multicolumn{2}{l}{Number of Parameters [Medium\,>\,Small]} & 1.65 & 0.18 & 9.15 & <\,.001 \\
\multicolumn{2}{l}{Epistemic Certainty [Not Low\,>\,Low]} & 3.08 & 0.22 & 14.08 & <\,.001 \\
\multicolumn{2}{l}{Statement Type Fact [>\,Belief]} & 3.79 & 0.22 & 17.28 & <\,.001 \\
\multicolumn{2}{l}{Statement Type Agent 1 [>\,Agent 0]} & $-$0.38 & 0.22 & $-$1.75 & .081 \\
\multicolumn{2}{l}{Statement Type Current [>\,Previous]} & 0.39 & 0.25 & 1.59 & .113 \\
\multicolumn{2}{l}{QA Format Indirect Slot [>\,Direct Slot]} & $-$2.65 & 0.27 & $-$9.73 & <\,.001 \\
\multicolumn{2}{l}{QA Format Indirect Sentence [>\,Direct Slot]} & 3.99 & 0.27 & 14.58 & <\,.001 \\
\multicolumn{2}{l}{Number of Parameters $\times$ Epistemic Certainty} & $-$2.99 & 0.44 & $-$6.85 & <\,.001 \\
\multicolumn{2}{l}{Number of Parameters $\times$ Statement Type Fact} & $-$1.01 & 0.44 & $-$2.31 & .021 \\
\multicolumn{2}{l}{Number of Parameters $\times$ Statement Type Current} & $-$1.22 & 0.49 & $-$2.48 & .013 \\
\multicolumn{2}{l}{Number of Parameters $\times$ QA Format Indirect Slot} & 2.01 & 0.54 & 3.70 & <\,.001 \\
\multicolumn{2}{l}{Number of Parameters $\times$ QA Format Indirect Sentence} & $-$1.53 & 0.55 & $-$2.81 & .005 \\
\midrule
Observations & 1440 & & & \\
$R^2_{\mathrm{\,Tjur}}$ & .597 & & & \\
AIC & 1009.2 & & & \\
\bottomrule
\end{tabular}
\caption{Logistic regression results on \textbf{Qwen2.5-7B/72B} data from Experiment 2, derived from the optimal regression model selected by AIC.}
\label{table: exp2-Qwen2.5}
\end{table*}

\begin{table*}[h]
\small
\centering
\begin{tabular}{llrcrr}
\toprule
& & \multicolumn{4}{c}{\textbf{Correct Response}} \\
\multicolumn{2}{l}{Predictor} & $b~~~$ & $SE$ & $z~~~$ & $p~~~~~$ \\
\midrule
\multicolumn{2}{l}{(Intercept)} & $-$0.20 & 0.09 & $-$2.11 & .035 \\
\multicolumn{2}{l}{Number of Parameters [Medium\,>\,Small]} & 1.88 & 0.19 & 9.87 & <\,.001 \\
\multicolumn{2}{l}{Epistemic Certainty [Not Low\,>\,Low]} & 1.36 & 0.32 & 4.31 & <\,.001 \\
\multicolumn{2}{l}{Statement Type Fact [>\,Belief]} & 4.79 & 0.34 & 14.07 & <\,.001 \\
\multicolumn{2}{l}{Statement Type Agent 1 [>\,Agent 0]} & 0.21 & 0.22 & 0.97 & .331 \\
\multicolumn{2}{l}{Statement Type Current [>\,Previous]} & $-$0.22 & 0.22 & $-$0.99 & .325 \\
\multicolumn{2}{l}{QA Format Indirect Slot [>\,Direct Slot]} & $-$1.76 & 0.23 & $-$7.81 & <\,.001 \\
\multicolumn{2}{l}{QA Format Indirect Sentence [>\,Direct Slot]} & 1.26 & 0.22 & 5.60 & <\,.001 \\
\multicolumn{2}{l}{Number of Parameters $\times$ Statement Type Fact} & $-$4.50 & 0.67 & $-$6.73 & <\,.001 \\
\multicolumn{2}{l}{Number of Parameters $\times$ Epistemic Certainty} & $-$5.30 & 0.64 & $-$8.34 & <\,.001 \\
\midrule
Observations & 1440 & & & \\
$R^2_{\mathrm{\,Tjur}}$ & .546 & & & \\
AIC & 1064.9 & & & \\
\bottomrule
\end{tabular}
\caption{Logistic regression results on \textbf{Llama3-8B/70B} data from Experiment 2, derived from the optimal regression model selected by AIC.}
\label{table: exp2-Llama3}
\end{table*}

\begin{table*}[h]
\small
\centering
\begin{tabular}{llrcrr}
\toprule
& & \multicolumn{4}{c}{\textbf{Correct Response}} \\
\multicolumn{2}{l}{Predictor} & $b~~~$ & $SE$ & $z~~~$ & $p~~~~~$ \\
\midrule
\multicolumn{2}{l}{(Intercept)} & 0.59 & 0.09 & 6.32 & <\,.001 \\
\multicolumn{2}{l}{Number of Parameters [Medium\,>\,Small]} & 2.70 & 0.19 & 14.57 & <\,.001 \\
\multicolumn{2}{l}{Epistemic Certainty [Not Low\,>\,Low]} & $-$0.55 & 0.17 & $-$3.26 & .001 \\
\multicolumn{2}{l}{Statement Type Fact [>\,Belief]} & 2.74 & 0.19 & 14.73 & <\,.001 \\
\multicolumn{2}{l}{Statement Type Agent 1 [>\,Agent 0]} & 0.43 & 0.20 & 2.18 & .029 \\
\multicolumn{2}{l}{Statement Type Current [>\,Previous]} & 0.19 & 0.22 & 0.87 & .386 \\
\multicolumn{2}{l}{QA Format Indirect Slot [>\,Direct Slot]} & $-$0.46 & 0.20 & $-$2.23 & .026 \\
\multicolumn{2}{l}{QA Format Indirect Sentence [>\,Direct Slot]} & 1.06 & 0.21 & 5.00 & <\,.001 \\
\multicolumn{2}{l}{Number of Parameters $\times$ Epistemic Certainty} & $-$4.78 & 0.34 & $-$14.24 & <\,.001 \\
\multicolumn{2}{l}{Number of Parameters $\times$ Statement Type Fact} & 2.28 & 0.37 & 6.14 & <\,.001 \\
\multicolumn{2}{l}{Number of Parameters $\times$ Statement Type Agent 1} & 1.17 & 0.40 & 2.97 & .003 \\
\multicolumn{2}{l}{Number of Parameters $\times$ QA Format Indirect Sentence} & $-$1.22 & 0.37 & $-$3.27 & .001 \\
\midrule
Observations & 1440 & & & \\
$R^2_{\mathrm{\,Tjur}}$ & .465 & & & \\
AIC & 1217.9 & & & \\
\bottomrule
\end{tabular}
\caption{Logistic regression results on \textbf{Llama3.1-8B/70B} data from Experiment 2, derived from the optimal regression model selected by AIC.}
\label{table: exp2-Llama3.1}
\end{table*}

\clearpage
\onecolumn
\section{ROC Curves for Logistic Regression Models}
\label{app:roccurves}

\subsection{Experiment 1}
\label{app:roccurves:exp1}

%\clearpage
\subsection{Experiment 2}
\label{app:roccurves:exp2}

\twocolumn

% linguistic diversity

\end{document}